%% file: main.tex
\documentclass[10pt,twocolumn,letterpaper]{article}

\usepackage[pagenumbers]{wacv} %

\usepackage{kotex}
\usepackage{amsmath, amssymb}
\usepackage{algorithm}
\usepackage{algpseudocode}
\usepackage{graphicx}
\usepackage{multirow}
\usepackage{booktabs}
\usepackage{caption}
\usepackage{pgfplots}
\usepackage[hang,flushmargin]{footmisc}
\usepackage[accsupp]{axessibility}  %

\input{macro}

\definecolor{wacvblue}{rgb}{0.21,0.49,0.74}
\usepackage[pagebackref,breaklinks,colorlinks,allcolors=wacvblue]{hyperref}

\def\confName{WACV}
\def\confYear{2026}

\title{Edge-Aware Image Manipulation via Diffusion Models\\with a Novel Structure-Preservation Loss}

\author{
    Minsu Gong$^{1, \ast, \dagger}$ \and
    Nuri Ryu$^{2, \ast}$ \and
    Jungseul Ok$^{2}$ \and
    Sunghyun Cho$^{2}$ \\
    \vspace{-4.5mm}
    \and
    $^1$Planby Technologies \hspace{10mm} $^{2}$POSTECH
}

\begin{document}
\maketitle
{\let\thefootnote\relax\footnotetext{
$^{\ast}$ Equal contribution. \\
$^{\dagger}$ Work done while the author was at POSTECH. \\
}}
\input{sec/0_abstract}
\notoc{\input{sec/1_intro}}
\notoc{\input{sec/2_related}}
\notoc{\input{sec/3_method}}
\notoc{\input{sec/4_experiment}}
\notoc{\input{sec/5_conclusion}}
{
    \small
    \bibliographystyle{ieeenat_fullname}
    \bibliography{main}
}

\input{sec/X_suppl}

\end{document}

%% file: macro.tex
\def\LossName{SPL}

\newcommand{\notoc}[1]{%
    \addtocontents{toc}{\protect\setcounter{tocdepth}{-1}}%
    #1%
    \addtocontents{toc}{\protect\setcounter{tocdepth}{2}}%
}

%% file: sec/0_abstract.tex
\begin{abstract}
Recent advances in image editing leverage latent diffusion models (LDMs) for versatile, text-prompt-driven edits across diverse tasks.
Yet, maintaining pixel-level edge structures—crucial for tasks such as photorealistic style transfer or image tone adjustment—remains as a challenge for latent-diffusion-based editing. 
To overcome this limitation, we propose a novel Structure Preservation Loss (SPL) that leverages local linear models to quantify structural differences between input and edited images.
Our training-free approach integrates SPL directly into the diffusion model's generative process to ensure structural fidelity. This core mechanism is complemented by a post-processing step to mitigate LDM decoding distortions, a masking strategy for precise edit localization, and a color preservation loss to preserve hues in unedited areas.
Experiments confirm SPL enhances structural fidelity, delivering state-of-the-art performance in latent-diffusion-based image editing.
Our code will be publicly released at \url{https://github.com/gongms00/SPL}.
\end{abstract}

%% file: sec/1_intro.tex
\section{Introduction}
Image editing has seen remarkable progress, yet developing a universal image editing method that preserves \textit{pixel-level edge structures} of the input image remains a long-standing challenge. 
By pixel-level edge structure, we mean the fine-grained spatial discontinuities in intensity that define object contours and texture details. 
Maintaining these structures is crucial for various tasks like relighting, tone adjustment, image harmonization, photorealistic style transfer, time-lapse generation, seasonal or weather changes, and background replacement. 
In these editing tasks, even minor structural distortions in the edited image, relative to the input, can compromise the intention of the edit.

Traditionally, pixel-level-edge-preserving image editing tasks have been approached individually using tailored methods~\cite{Deep_Single-Image_Portrait_Relighting_Zhou_2019_ICCV, Color_transfer_between_images_946629, Using_Color_Compatibility_for_Assessing_Image_Realism_4409107, Data-driven_hallucination_10.1145/2508363.2508419, Transient_Attributes_10.1145/2601097.2601101}.
While these methods achieve high structural fidelity, they face two significant drawbacks. 
First, none of these methods offer a unified framework that can handle multiple editing tasks, which results in task-specific pipelines with limited generalization.
Second, lacking a generative prior, they struggle to introduce additional details or creative variations aligned with desired edits.

Recent image editing methods leverage robust generative priors such as large-scale text-to-image latent diffusion models (LDMs)~\cite{stable_diffusion, sdxl}. This has enabled general image editing through text-based instructions, allowing them to address multiple tasks within a single framework~\cite{InstructPix2Pix_Brooks_2023_CVPR, P2P_hertz2023prompttoprompt, MasaCtrl_Cao_2023_ICCV, Imagic_Kawar_2023_CVPR, SDEdit_meng2022sdedit, InfEdit_Xu_2024_CVPR}.
While these LDM-based editing methods overcome the limitations of generalization and detail generation in traditional image editing methods, they often fall short of preserving the pixel-level structure of the input image even when strict preservation is needed. 
This challenge mainly stems from the absence of explicit guidance to maintain pixel-level edge structures.
Moreover, LDM-based editing methods also suffer from structural distortions arising from the lossy RGB-to-latent mapping, exacerbating structure preservation difficulties.

\input{figures/SPL_teaser}

To tackle these challenges, we propose \textit{Structure Preservation Loss} (\LossName{}), a novel loss function designed to enforce pixel-level edge structure fidelity between the input and edited output. 
The key idea behind the structure preservation loss is to leverage the notion of local linear models (LLMs) in image processing ~\cite{Guided_Filter, levin2007closed, zomet2002multi, he2010single} to capture and maintain structural relationships between images.
We incorporate the structure preservation loss into the sampling process of diffusion models, achieving high structure fidelity to the input image without harming its editing capabilities as demonstrated in ~\cref{fig:teaser}.
Our approach is plug-and-play and does not require any additional training on the base LDM-based editing model.
Furthermore, the modularity of our loss offers users fine-grained control to tailor structural constraints to their specific intent.
We also introduce a post-processing step that refines the decoded output to correct structural distortions arising from the latent-to-RGB mapping, ensuring pixel-level edge structure preservation.
Additionally, we propose a simple masking strategy that generates a high-resolution edit mask based on the edit text prompt to enable precise edit localization.
Our method establishes a general framework for structure-preserving editing, unifying a class of tasks traditionally handled by separate, specialized methods. By integrating our loss, conventional LDM-based editors can now overcome their inherent limitation in preserving pixel-level structures, significantly expanding their applicability and making them more versatile, all-purpose editing tools.

Experiments conducted on various editing tasks confirm that our method successfully produces edited images that maintain the input image's pixel-level edge structure while being faithful to the provided edit prompts.

To summarize, our main contributions are as follows:
\begin{itemize}
    \item We introduce a novel structure preservation loss that maintains the pixel-level edge structure of the input image during image editing.
    \item We propose a training-free latent-diffusion-model-based image editing method that allows pixel-level edge structure preservation while leveraging latent-diffusion models' generative image editing capabilities.
    \item We suggest a simple strategy to acquire a high-resolution edit mask from the diffusion model's internal features to support local editing.
    \item We demonstrate state-of-the-art quality in pixel-level structure-preserving image editing through comprehensive experiments.
\end{itemize}

%% file: figures/SPL_teaser.tex
\begin{figure}[t!]
\begin{center}
\includegraphics[width=\linewidth]{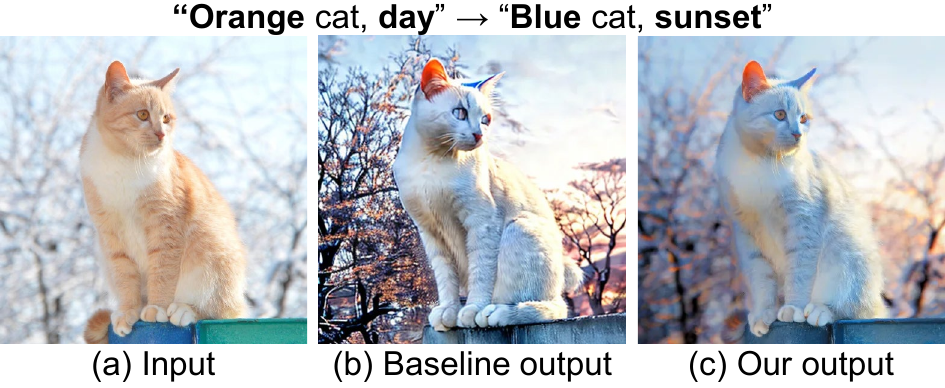}
\end{center}
\vspace{-2.5mm}
\caption{
\textbf{Structure-preserving edit.}
Our method achieves pixel-level structural fidelity without compromising the intended edit.
}
\label{fig:teaser}
\vspace{-4.0mm}
\end{figure}

%% file: sec/2_related.tex
\section{Related Work}

\paragraph{Structure-Preserving Image Editing}
Prior to the emergence of diffusion models, structure-preserving image editing tasks were predominantly tackled using specialized, task-dependent methods. Techniques for image relighting leveraged physics-based priors~\cite{Acquiring_the_reflectance_field_of_a_human_face_10.1145/344779.344855, Performance_relighting_and_reflectance_wenger2005performance, Deep_Single-Image_Portrait_Relighting_Zhou_2019_ICCV, Total_Relighting_10.1145/3450626.3459872, SwitchLight_kim2024switchlight, sun2019single, nestmeyer2020learning}, while tone adjustment relied on color transformations or look-up tables~\cite{chai2020supervised, gharbi2017deep, yan2016automatic, he2020conditional, kim2021representative, wang2021real, zeng2020learning, CLIPtone_lee2024cliptone}. Harmonization and background replacement employed image gradient or color statistics manipulation~\cite{jia2006drag, 10.1145/1201775.882269, sunkavalli2010multi, tao2013error, 10.1145/1179352.1141933, pitie2005n, reinhard2001color, xue2012understanding, ling2021region, PCA-Based_Knowledge_Distillation_Chiu_2022_CVPR}, and style transfer utilized image statistics or deep feature matching~\cite{reinhard2001color, pitie2005n, gatys2016image, li2016combining, simonyan2014very, Deep_Photo_Style_Transfer_Luan_2017_CVPR, yoo2019photorealistic, li2018closed, PCA-Based_Knowledge_Distillation_Chiu_2022_CVPR}.  
While these task-specific methods often demonstrated high structural fidelity within their respective domains, they are inherently limited in scope and generalization due to their domain-specific assumptions.

\paragraph{Diffusion-Based Image Editing}
Large-scale text-to-image LDMs~\cite{stable_diffusion, sdxl} have enabled versatile image editing methods, offering text-driven edits with a single pre-trained model \cite{Imagic_Kawar_2023_CVPR, SINE_Bao_2023_CVPR, DiffusionCLIP_Kim_2022_CVPR, InstructPix2Pix_Brooks_2023_CVPR, InstructDiffusion_Geng_2024_CVPR, HIVE_Zhang_2024_CVPR, SDEdit_meng2022sdedit, DiffEdit_couairon2023diffedit, Blended_Diffusion_Avrahami_2022_CVPR, Null-text_Mokady_2023_CVPR, EDICT_Wallace_2023_CVPR, PnPInversion_ju2024pnp, DDPMInversion_Huberman-Spiegelglas_2024_CVPR, P2P_hertz2023prompttoprompt, PNP_Tumanyan_2023_CVPR, Pix2pix-zero, MasaCtrl_Cao_2023_ICCV, InfEdit_Xu_2024_CVPR}. 
These methods broadly fall into training-based and training-free approaches, both facing challenges in preserving fine-grained structural details from the input image.

Training-based approaches include methods that overfit to a single image \cite{Imagic_Kawar_2023_CVPR, SINE_Bao_2023_CVPR} or learn from editing datasets \cite{DiffusionCLIP_Kim_2022_CVPR, InstructPix2Pix_Brooks_2023_CVPR, InstructDiffusion_Geng_2024_CVPR, HIVE_Zhang_2024_CVPR}. 
However, overfitting to a single image incurs a significant computational overhead, limiting its scalability, while dataset-trained models are limited by dataset biases. Furthermore, the training objectives of training-based methods do not guarantee pixel-level fidelity.

Training-free methods leverage a pre-trained diffusion model without additional training or fine-tuning. Early approaches edit images by steering the generative process of diffusion models to match the edit prompt by modifying the intermediate noisy latent representations~\cite{SDEdit_meng2022sdedit, DiffEdit_couairon2023diffedit, Blended_Diffusion_Avrahami_2022_CVPR}. However, these techniques are often limited in editing performance and struggle to retain the original image’s structure.
More recent training-free methods have improved on fidelity by manipulating the diffusion model’s denoising U-Net attention maps~\cite{P2P_hertz2023prompttoprompt, PNP_Tumanyan_2023_CVPR, Pix2pix-zero, MasaCtrl_Cao_2023_ICCV, InfEdit_Xu_2024_CVPR}. For example, Prompt-to-Prompt~\cite{P2P_hertz2023prompttoprompt} reuses cross-attention maps generated during input image generation to maintain its global layout, while Plug-and-Play~\cite{PNP_Tumanyan_2023_CVPR} injects self-attention to preserve local structure. Unifying these ideas, InfEdit~\cite{InfEdit_Xu_2024_CVPR} combines controls for both attention types. However, lacking an explicit pixel-level guidance, these methods only allow for coarse structure preservation.

\paragraph{Full-Reference Metrics for Structural Fidelity}
While full-reference metrics are often used as loss functions to preserve fidelity to a reference, they are ill-suited for structure-preserving editing tasks where appearance (e.g., color, brightness, contrast) is intentionally modified. These metrics typically fail to disentangle structural content from appearance attributes. For instance, pixel-wise losses like L1 and MSE penalize valid appearance changes, while perceptual metrics like LPIPS~\cite{LPIPS} do not explicitly enforce structural accuracy.
SSIM~\cite{SSIM} does account for structure yet remains sensitive to changes in brightness and contrast, as it evaluates luminance and contrast alongside structural correlation.

%% file: sec/3_method.tex
\input{figures/SPL_diagram}

\section{Method}
Our method performs LDM-based image editing while preserving the pixel-level edge structure of the input. To achieve this, we first introduce a novel structure preservation loss that explicitly quantifies structural discrepancies (\cref{subsec:SPL}). We then integrate SPL into the LDM's denoising pipeline  (\cref{subsec:structure_preserving_ldm}).
Additionally, to offer users precise control over which regions should remain unchanged, we also propose a text-driven edit mask generation scheme to be used in conjunction with the structure preservation loss (\cref{subsec:cross_attention_mask}).

\subsection{Structure Preservation Loss}
\label{subsec:SPL}
\paragraph{Local Linear Model for Structure Preservation}
Our structure preservation loss is designed to quantify structural differences between a source image $I^S$ and an edited output $I^E$. 
To achieve this, our loss is grounded in the local linear model, which is well-established for its structure awareness and has been applied to various tasks that require accurate structure preservation, e.g., alpha matting, edge-aware filtering, dehazing, joint upsampling, and super-resolution~\cite{Guided_Filter, levin2007closed, zomet2002multi, he2010single}. 
The key assumption is that for structure to be preserved, the edited image $I^E$ must be a local linear transformation of the source image $I^S$.
We visualize this idea in \cref{fig:SPL_diagram}-1.

We formalize this relationship mathematically within a local window $\omega_k$ as:
\begin{equation}
    I^S_i = a_k \, I^E_i + b_k, \quad \forall i \in \omega_k,
    \label{eq:local_linear_model}
\end{equation}
where \(a_k\) and \(b_k\) are coefficients assumed to be constant within the window centered at pixel $k$. 
A key property of this model is that it preserves edges, since taking the gradient of \cref{eq:local_linear_model} yields $\nabla I^S_i = a_k \nabla I^E_i$. 
The coefficients are obtained by solving a least-squares problem, yielding the closed-form solution:
\begin{align}
    a_k =& \frac{\frac{1}{|\omega_k|} \sum_{i \in \omega_k} I^E_i I^S_i - \mu^E_k \mu^S_k}{\left(\sigma^E_k\right)^2 + \rho},~~~~\textrm{and}
    \label{eq:a_coefficient} \\
    b_k =& \mu^S_k - a_k \mu^E_k,
    \label{eq:b_coefficient}
\end{align}
where \(\mu_k^E\) and \(\mu_k^S\) are the mean intensities of \(I^E\) and \(I^S\) in \(\omega_k\), respectively. \(\sigma^E_k\) is the standard deviation of \(I^E\) in $\omega_k$, and \(|\omega_k|\) is the number of pixels in the window.
$\rho$ is a small constant to avoid degenerate solutions.
We provide a detailed derivation in the supplementary.

While previous works employ the local linear model as a filtering mechanism, our key contribution is to adapt it as a differentiable loss. This loss serves as an explicit structural guide for the diffusion model's generative process, ensuring fidelity across a wide range of editing tasks.

\paragraph{Structure Preservation Loss}
Based on \cref{eq:local_linear_model}, we define the directional structure difference between $I^E$ and $I^S$ as:
\begin{equation}
    D_k(I^E, I^S) = \frac{1}{|\omega_k|} \sum_{i \in \omega_k} \left( I_i^{E \to S} - I_i^S \right)^2, 
    \label{eq:directional_difference_ab}
\end{equation}
where $I_i^{E \to S}$ represents a linearly transformed local patch of $I^E$, defined as $I_i^{E \to S} = a_k I_i^E + b_k$.
Here, $D_k(I^E, I^S)$ quantifies the discrepancy between the transformed and original images. It is noteworthy that $D_k(I^E, I^S) \neq D_k(I^E, I^S)$, as the operation is not symmetric. Consequently, $D_k(I^E, I^S)$ alone may fail to fully capture the structural differences between two images.
For example, consider a case where a region in $I^S$ is structurally flat while the corresponding region in $I^E$ contains structural details (see highlighted region in \cref{fig:SPL_diagram}-2). 
In such a scenario, a trivial transformation ($a_k \approx 0, b_k \approx \mu_k^S$) can map $I^E$ to $I^S$. 
This results in a deceptively low value for $D_k(I^E, I^S)$ despite the significant structural disparities, as shown in the corresponding error map (\cref{fig:SPL_diagram}-2-c).
To address this limitation, we introduce a reverse mapping $D_k(I^S, I^E)$ (\cref{fig:SPL_diagram}-2-d).
Finally, we combine the forward and reverse mappings to define the structure preservation loss:
\begin{equation} \mathcal{L}_{\text{SPL}}(I^E, I^S) = \frac{1}{N} \sum_{k} \left( D_k(I^E, I^S) + D_k(I^S, I^E) \right), \label{eq:structure_preservation_loss} \end{equation}
where $N$ is the total number of pixels in the image, and the summation is performed over all pixels $k$. By enforcing consistency across both forward ($E \to S$) and reverse ($S \to E$) transformations, this loss function robustly penalizes mismatches in underlying structures.
For RGB images, we convert them to HSI color space, and use the I channel to compute $\mathcal{L}_\textrm{SPL}$ in our approach.

\paragraph{Color Preservation Loss}
In tasks like local color editing, we must ensure that changing an object's color does not alter the hues of its surroundings. To address this, we introduce an optional color preservation loss, \(\mathcal{L}_{\text{CPL}}\), that can be used alongside \(\mathcal{L}_{\text{SPL}}\). 
We compute \(\mathcal{L}_{\text{CPL}}\) as the mean squared error between the Cb and Cr channels of \(I^A\) and \(I^B\) in the YCbCr color space:
\begin{equation}
    \mathcal{L}_{\text{CPL}} (I^E, I^S) = \frac{1}{N} \sum_{k} \left\| \text{CbCr}(I^E_k) - \text{CbCr}(I^S_k) \right\|_2^2,
    \label{eq:color_preservation_loss}
\end{equation}

We empirically found that using HSI for $\mathcal{L}_\textrm{SPL}$ and YCbCr for $\mathcal{L}_\textrm{CPL}$ yields higher-quality results.
This is because they provide simple linear conversions from the RGB color space. This simplicity enables more efficient loss optimization. Furthermore, the I channel in the HSI color space, calculated as the average of the color channels with equal weights, ensures uniform updates across all RGB components when optimizing $\mathcal{L}_\textrm{SPL}$.

\input{figures/pipeline_fig}

\subsection{Structure-Preserving Editing with LDMs}
\label{subsec:structure_preserving_ldm}
In this section, we explain how our structure preservation loss enhances the denoising process of LDMs to achieve pixel-level structure-preserving image editing. We first outline conventional coarse-structure-preserving editing in LDMs and its limitations, then describe our approach, which integrates \(\mathcal{L}_{\text{SPL}}\) through the denoising diffusion process and a post-processing step. We provide an overview in~\cref{fig:pipeline}.

\paragraph{Review of Coarse-Structure-Preserving Image Editing in LDMs}
In LDMs, an encoder \(\mathcal{E}\) maps images to latents, and a decoder \(\mathcal{D}\) reconstructs images from these latents.
The diffusion process of LDMs is defined in this latent space, enabling efficient image generation and manipulation~\cite{sdxl, stable_diffusion}.

Coarse-structure-preserving image editing with LDMs, such as Prompt-to-Prompt~\cite{P2P_hertz2023prompttoprompt} and InfEdit~\cite{InfEdit_Xu_2024_CVPR}, typically involves three inputs: a source image \(I_{\text{src}}\), its associated text prompt \(p_{\text{src}}\), and a target edit prompt \(p_{\text{edit}}\). %
The editing process begins with a latent code \(z_T\) sampled from pure Gaussian noise, i.e., \(z_T \sim \mathcal{N}(0, I)\), where \(T\) is the maximum timestep.
The reverse diffusion process then iteratively denoises this latent from \(t = T\) to \(t = 1\), guided by a noise prediction model \(\epsilon_\theta\), parameterized by \(\theta\). 

To achieve coarse-structure-preserving edits, \(\epsilon_\theta\) is conditioned on the target prompt \(p_{\text{edit}}\) and source features \(f^{\text{src}}_t\) (e.g., attention maps, estimated noise), extracted from the generation of the source image  \(I_{\text{src}}\) using its prompt \(p_{\text{src}}\)~\cite{P2P_hertz2023prompttoprompt, InfEdit_Xu_2024_CVPR}. 
This conditioning preserves the coarse structure of the source image during the generation of the edited image. 
For a forward process defined as $z_t = \alpha_t z_0 + \sigma_t \epsilon$ with $\epsilon \sim \mathcal{N}(0,I)$, the predicted denoised latent $\hat{z}_0^{(t)}$ at each timestep $t$ is computed as:
\begin{equation}
    \hat{z}_0^{(t)} = \frac{1}{a_t} \left( z_t - b_t \epsilon_\theta\left(z_t, t, p_{\text{edit}}, f^{\text{src}}_t\right) \right).
    \label{eq:z0_pred}
\end{equation}
From $\hat{z}_0^{(t)}$, the noisy latent for $t-1$ is computed as:
\begin{equation}
    z_{t-1} = \mathcal{S}\left(\hat{z}_0^{(t)}, z_t, t, \hat{\epsilon}_t \right),
    \label{eq:denoising_update}
\end{equation}
where \(\mathcal{S}\) denotes a sampling function and \(\hat{\epsilon}_t\) is the estimated noise \(\epsilon_\theta\left(z_t, t, p_{\text{edit}}, f^{\text{src}}_t\right)\).
The iterative denoising process then advances to timestep $t-1$.
While the attention conditioning strategy leveraging $f_t^\textrm{src}$ in \cref{eq:z0_pred} preserves coarse image layouts, it does not retain fine-grained structural details of the input image, as will be shown in \cref{sec:experiment}.

\paragraph{Editing with Structure Preservation Loss}
Our task of preserving pixel-level edge structure is fundamentally defined in the image space and thus requires a corresponding pixel-space guidance. Hence, we apply \(\mathcal{L}_{\text{SPL}}\) to the VAE-decoded image of the denoised latent $\hat{z}_0^{(t)}$ at intermediate timesteps, following prior works that have successfully guided latent denoising with pixel-space objectives~\cite{lee2023syncdiffusion, rout2023solving_PLSD, zilberstein2025repulsive_RLSD, Xinqi2024DiffBIR}.

At each timestep $t$, we compute $\hat{z}_0^{(t)}$ using \cref{eq:z0_pred}.
Next, we update $\hat{z}_0^{(t)}$ by solving:
\begin{equation}
\tilde{z}=\mathop{\mathrm{argmin}}_{\hat{z}}\mathcal{L}_\textrm{SPL}\left(I_\textrm{src},\mathcal{D}(\hat{z})\right)+\lambda\mathcal{L}_\textrm{CPL}\left(I_\textrm{src},\mathcal{D}(\hat{z})\right)
\label{eq:diffusion_with_SPL}
\end{equation}
where $\hat{z}$ is initially set to $\hat{z}_0^{(t)}$, $\tilde{z}$ is the updated latent, and $\lambda$ is a weight for the optional color preservation loss, which we set to a small value.
Finally, $\tilde{z}$ is substituted into \cref{eq:denoising_update} in place of $\hat{z}_0^{(t)}$, and the denoising process continues to timestep $t-1$.
The updated latent $\tilde{z}$ guides the subsequent denoising steps to not only retain the source image's structural details, but also synthesize natural-looking image contents that align with the retained details.

To solve \cref{eq:diffusion_with_SPL}, we adopt the gradient descent method. However, direct application of gradient descent requires repeated, computationally expensive gradient calculations due to backpropagation through $\mathcal{D}$. To circumvent this problem, we introduce an auxiliary image $\hat{I}$, initialized as $\hat{I}=\mathcal{D}(\hat{z}_0^{(t)})$, and reformulate the optimization problem as:
\begin{equation}
\tilde{I}=\mathop{\mathrm{argmin}}_{\hat{I}}\mathcal{L}_\textrm{SPL}\left(I_\textrm{src},\hat{I}\right)+\lambda\mathcal{L}_\textrm{CPL}\left(I_\textrm{src},\hat{I}\right).
\label{eq:diffusion_with_SPL2}
\end{equation}
After solving this, we compute $\tilde{z}=\mathcal{E}(\tilde{I})$ and substitute it into \cref{eq:denoising_update}.
\cref{eq:diffusion_with_SPL2} is efficiently solved using a gradient descent update:
\begin{equation}
    \hat{I} \leftarrow \hat{I} - \eta \nabla_{\hat{I}}\left\{ \mathcal{L}_{\text{SPL}}(I_{\text{src}}, \hat{I})+\lambda\mathcal{L}_{\text{CPL}}(I_{\text{src}}, \hat{I})\right\},
\end{equation}
where $\eta$ is the learning rate.

We apply this optimization selectively during the later stages of the denoising process (i.e., timesteps \(t \leq t_{\text{SPL}}\)). This approach allows for early denoising steps to maintain generative flexibility while later stages prioritize accurate preservation of structural details. Further analysis of this scheduling is provided in the supplementary.

\paragraph{Post-Processing Step}
While the diffusion process with the structure preservation loss improves structural consistency, the final decoding \(\hat{I}_0 = \mathcal{D}(\hat{z}_0)\) may still lose fine details due to its inherent compression characteristics. To mitigate this, we introduce a post-processing step that refines \(\hat{I}_0\) directly in the image space by solving \cref{eq:diffusion_with_SPL2}.
This step ensures pixel-level fidelity to the source image's fine structures, compensating for the decoder's limitations.

\input{figures/mask_fig}

\subsection{Cross-Attention Mask Upsampling for Structure-Preserving Localized Editing}
\label{subsec:cross_attention_mask}

For localized editing such as background replacement (\cref{fig:mask}), we apply structure and color preservation losses to specific regions defined by a binary mask. We derive an initial coarse mask by thresholding the low-resolution (\(16 \times 16\)) cross-attention map corresponding to the edit prompt \cite{P2P_hertz2023prompttoprompt}  (\cref{fig:mask}-c). However, naively upsampling this mask with bilinear interpolation produces blurry and misaligned boundaries.
To overcome this, we introduce a simple yet effective iterative refinement algorithm. Starting with the coarse mask, we progressively upscale it by a factor of two. At each step, we use the input image, downscaled to the current mask's resolution, as a guidance image for the guided filter~\cite{Guided_Filter}. This process sharpens and aligns the mask's boundaries by transferring structural details from the guide image. The result is a high-resolution mask with sharp, structurally accurate boundaries (\cref{fig:mask}-d), enabling precise and artifact-free local editing (\cref{fig:mask}-e). We provide a pseudocode and further details in the supplementary.

%% file: figures/SPL_diagram.tex
\begin{figure*}[t!] 
\begin{center}
\includegraphics[width=\linewidth]{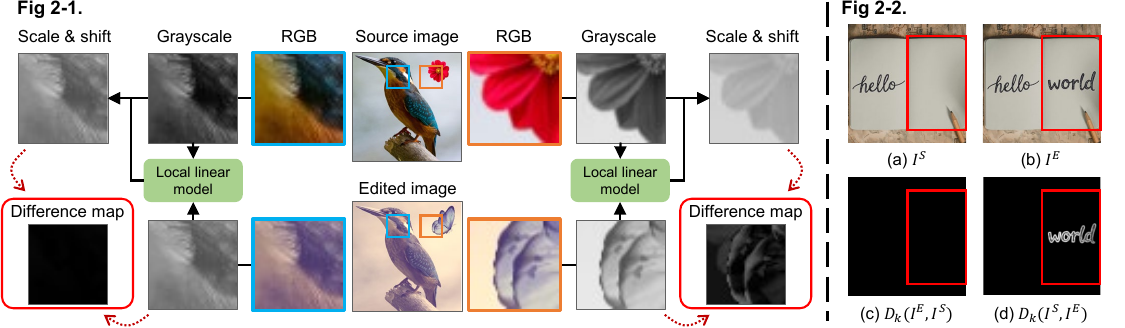}
\end{center}
\vspace{-3mm}
\caption{
\textbf{Motivation of the structure preservation loss}. \textbf{(Fig 2-1)}
An edited image may contain both structure-preserving (marked in blue) and structure-breaking regions (marked in orange). Our approach is motivated by the local linear model's ability to analyze these regions on a local window-by-window basis. (Left) When structure is preserved, the model finds an accurate linear fit, resulting in low error. (Right) When structure is broken, the model fails to find a good fit, producing a high error that signals the distortion.
\textbf{(Fig 2-2)} Unidirectional structural comparison fails to fully capture mutual differences, motivating the bidirectional design of our structure preservation loss.
}
\vspace{-2.0mm}
\label{fig:SPL_diagram}
\end{figure*}

%% file: figures/pipeline_fig.tex
\begin{figure}[t!] 
\begin{center}
\includegraphics[width=\linewidth]{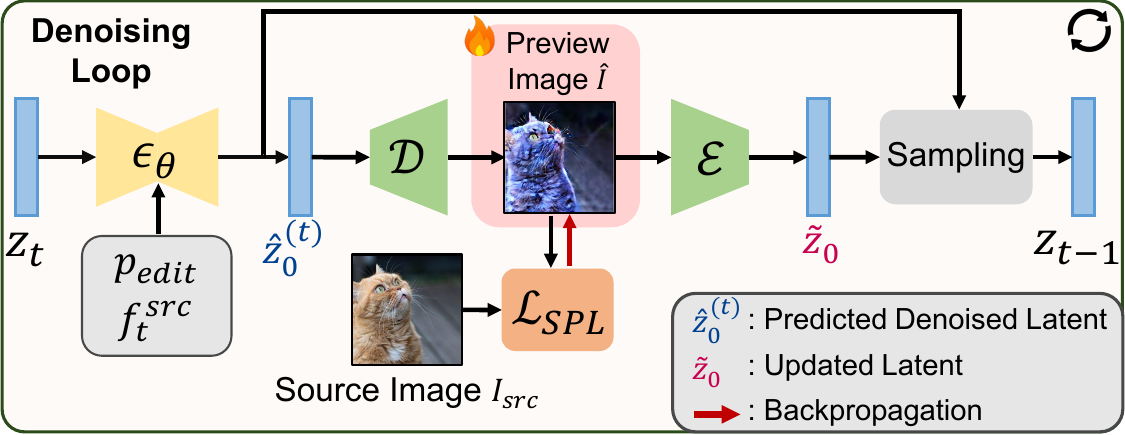}
\end{center}
\vspace{-3mm}
\caption{
\textbf{Structure-preserving denoising loop}. 
At each denoising timestep $t$, we decode the predicted clean latent $\hat{z}_0^{(t)}$ to compute our structure preservation loss $\mathcal{L}_{\text{SPL}}$ in the image space. The resulting gradient is then used to update the latent, producing a corrected version $\tilde{z}$ that steers the generation trajectory to maintain structural fidelity for the subsequent denoising step.
}
\vspace{-2.0mm}
\label{fig:pipeline}
\end{figure}

%% file: figures/mask_fig.tex
\begin{figure}[t!]
\begin{center}
\includegraphics[width=\linewidth]{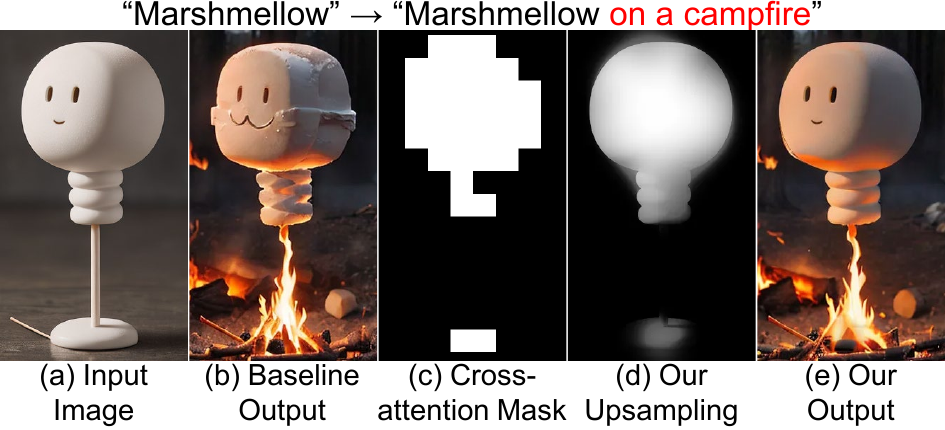}
\end{center}
\vspace{-3mm}
\caption{
\textbf{Cross-attention mask upsampling.}
For the background editing task, the baseline method  unintentionally distorts the foreground (b). By upsampling the coarse attention map (c), our method generates a sharp, high-resolution mask (d). This enables targeted application of our structure loss, preserving the foreground's fidelity in the final edit (e).
}
\label{fig:mask}
\vspace{-2.0mm}
\end{figure}

%% file: sec/4_experiment.tex
\input{figures/qual_global_color_change_fig}
\input{figures/qual_local_structure_change}

\section{Experiment}
\label{sec:experiment}
\paragraph{Implementation Details}
We integrate our method into InfEdit~\cite{InfEdit_Xu_2024_CVPR}, a training-free, latent-diffusion-based framework for image editing.
While InfEdit is our primary backbone, our approach is model-agnostic. We demonstrate its application to another backbone in the supplementary.
For tasks requiring localized edits, we use the color preservation loss and the upscaled edit masks as detailed in ~\cref{subsec:cross_attention_mask}. 
We report the hyperparameters in the supplementary. 

\subsection{Evaluation Details}
\label{ssec:ev_det}
\paragraph{Benchmark for Diffusion-Based Image Editing Models}
We primarily evaluate our method on a curated subset of PIE-Bench~\cite{PnPInversion_ju2024pnp}.
This curation is necessary because standard benchmarks, such as AnyEdit~\cite{Yu_2025_CVPR_AnyEdit} or RealEdit~\cite{Sushko_2025_CVPR_RealEdit}, often do not strictly distinguish edits requiring pixel-level edge preservation, frequently mixing them within broader categories. Our curated benchmark provides a more controlled setting for a targeted evaluation. We categorize the editing tasks based on the scope of structure preservation:
\begin{enumerate}[label=\arabic*)]
    \item \textbf{Global editing}: Edits applied across the entire image, requiring structure preservation of the whole input image. Examples include relighting, photorealistic style transfer, tone adjustment, and time-lapse generation.
    \item \textbf{Local editing}: Edits requiring structure preservation of specific regions. Examples include background replacement and weather changes, where only the structure of the foreground needs to be preserved.
\end{enumerate}
For this adapted subset, we generate task-specific textual prompts using GPT-4o~\cite{gpt4}, followed by manual refinement, resulting in approximately 470 image-prompt pairs.

We also provide a supplementary evaluation on the standard AnyEdit~\cite{Yu_2025_CVPR_AnyEdit} benchmark and show that the results confirm the findings from our main experiments.
We conduct our evaluation on three categories from AnyEdit that align with our focus on structure-preserving edits: tone transfer, color alteration, and background change.

\paragraph{Evaluation Metric} 

We assess our method based on two key criteria: 

\textbf{(1) Preservation}: 
We measure how well the edited image preserves the original image's features. 
Our evaluation relies on a suite of standard, independent metrics, including SSIM~\cite{SSIM} for structural similarity, LPIPS~\cite{LPIPS} for perceptual similarity, FSIM~\cite{5705575_FSIM} for feature similarity, and GMSD~\cite{xue2013gradient_GMSD} for gradient similarity. We also report our structure preservation loss as it offers a direct measure for the fine-grained structural fidelity that current metrics do not exclusively capture.

\textbf{(2) Prompt fidelity}:
We assess how well the edited image aligns with the edit text prompt. We employ CLIP~\cite{CLIP} Score to measure the similarity between the target edit prompt and the edited image, and the CLIP Directional Similarity to quantify consistency between the textual prompt changes and corresponding image edits.

\input{tables/diffusion_editing_quan}
\input{tables/supple_anyedit}

\subsection{Comparison with SoTA Editing Methods}

We compare our method against state-of-the-art latent diffusion-based editing methods, including InstructPix2Pix~\cite{InstructPix2Pix_Brooks_2023_CVPR}, InfEdit~\cite{InfEdit_Xu_2024_CVPR}, DDPM Inversion~\cite{DDPMInversion_Huberman-Spiegelglas_2024_CVPR}, GNRI~\cite{samuel2025lightningfast_GNRI} and a combination of null-text inversion~\cite{Null-text_Mokady_2023_CVPR} and Prompt-to-Prompt~\cite{P2P_hertz2023prompttoprompt}. InstructPix2Pix relies on supervised training from editing datasets, InfEdit uses unified attention control for text-guided edits, DDPM Inversion employs inversion-based editing, GNRI introduces a fast inversion technique for real-time editing. Prompt-to-Prompt shows a source image layout preserving editing, where the source image is inverted with null-text inversion.

Quantitative evaluations on both PIE-Bench (\cref{tab:diffusion_quan}) and the AnyEdit benchmark (\cref{tab:anyedit_quan}) indicate that our method significantly outperforms baselines in preservation.
This improvement is largely attributed to our structure preservation loss, achieved without compromising prompt fidelity.
Qualitative comparisons further illustrate these advantages. In global editing tasks (\cref{fig:qual_global_color_change}), InstructPix2Pix often generates either overly exaggerated or insufficient edits, DDPM Inversion and GNRI struggle with both structure preservation and prompt fidelity, and InfEdit preserves only coarse-level structure, compromising fine structural details. In contrast, our method accurately reflects edit prompts while precisely preserving pixel-level edge structures.
For local editing tasks (\cref{fig:qual_local_structure_change}), our proposed mask acquisition method effectively confines edits to regions specified by the prompt (e.g., weather or background), maintaining key elements that remain the same in both source and target prompts. Conversely, methods without region-specific editing inadvertently change the entire image, failing to respect localized changes described by the prompt.

In the supplementary, we also provide comparisons with task-specific structure-preserving methods, including photo-realistic style transfer, image harmonization, tone adjustment, seasonal change, and time-lapse editing, alongside qualitative results on the AnyEdit benchmark.

\subsection{Analysis}
\label{subsec:analysis}

\paragraph{Comparison with Other Structure-Similarity Losses}
We illustrate the effectiveness of our structure preservation loss compared to other common structural similarity measures, e.g., SSIM, MSE, and the directional structure difference loss in~\cref{eq:directional_difference_ab}.
The editing task transforms the input image (\cref{fig:ablation_loss}-a) into a ``futuristic night" style.
Since MSE and SSIM penalize pixel-wise luminance and contrast differences, using them leads to results that fail to fully reflect the intended night-style contrast and brightness (\cref{fig:ablation_loss}-b, c). 
The directional structure difference loss does not effectively penalize structural artifacts when the source image can be approximated by the edit image with a local linear transform, causing visible distortions at object boundaries (\cref{fig:ablation_loss}-d). 
In contrast, our full structure preservation loss successfully maintains pixel-level edge fidelity while accurately following the target edit prompt (\cref{fig:ablation_loss}-e).

\paragraph{Component-wise Ablation}
The ablation studies in~\cref{fig:ablation_SPL} and ~\cref{tab:ablation_component} show the contribution of each component in the structure preservation loss.
In ~\cref{fig:ablation_SPL}, we edit the color of the cover text from red to yellow. The baseline undesirably alters elements like clothing and cover text (\cref{fig:ablation_SPL}-b). Adding SPL first improves overall structural preservation (\cref{fig:ablation_SPL}-c). Then, CPL with a mask excluding the book specifically restores the original clothing color (\cref{fig:ablation_SPL}-d), and post-processing recovers fine details like the cover text (\cref{fig:ablation_SPL}-e).
Quantitative results in
~\cref{tab:ablation_component} also falls in line with the qualitative result, where each component builds up better preservation without loss of prompt fidelity.

We also show the necessity of incorporating the structure
preservation loss within the diffusion pipeline in~\cref{fig:ablation_PP}. Note that applying our post-processing step to the edited result obtained without this intermediate guidance is equivalent to applying SPL directly to the edited image (\cref{fig:ablation_PP}-b). However, this approach cannot prevent LDM from generating structural changes that go beyond what the post-processing step can handle (\cref{fig:ablation_PP}-c).

\paragraph{Computational Cost}
We analyze the computational cost of our method in~\cref{tab:computational}, with all experiments conducted on a single NVIDIA A6000 GPU. Our method is substantially faster than inversion-based approaches like DDPM Inversion~\cite{DDPMInversion_Huberman-Spiegelglas_2024_CVPR} and the combination of null-text inversion~\cite{Null-text_Mokady_2023_CVPR} and Prompt-to-Prompt~\cite{P2P_hertz2023prompttoprompt}. Compared to the baseline InfEdit~\cite{InfEdit_Xu_2024_CVPR}, our approach adds a modest overhead of approximately 2 seconds. We argue this is a valuable trade-off, as this overhead enables the precise, pixel-level structural control that the baseline lacks.

\input{figures/ablation_loss_fig}
\input{figures/ablation_SPL_component}
\input{tables/ablation_component}

\input{figures/ablation_only_pp}
\input{tables/computational_cost}

%% file: figures/qual_global_color_change_fig.tex
\begin{figure*}[t!] 
\begin{center}
\includegraphics[width=\linewidth]{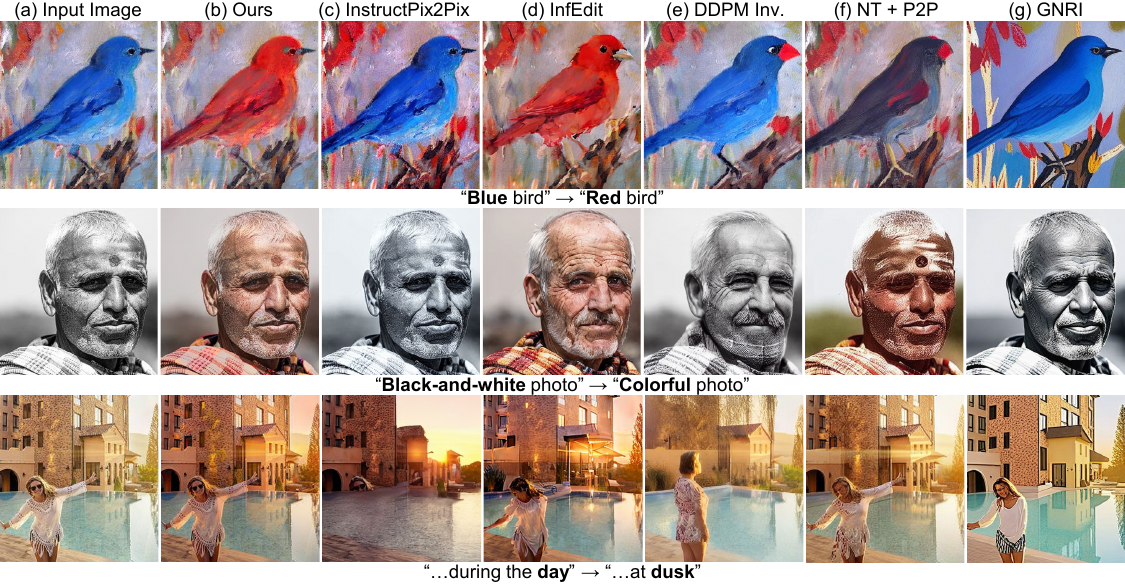}
\end{center}
\vspace{-3mm}
\caption{
\textbf{Qualitative comparison of global editing tasks.}
Our method (b) successfully applies the edit while preserving fine-grained structural details. Other methods (c-g) exhibit either low prompt fidelity or significant structural distortions
}
\label{fig:qual_global_color_change}
\end{figure*}

%% file: figures/qual_local_structure_change.tex
\begin{figure*}[t!] 
\begin{center}
\includegraphics[width=\linewidth]{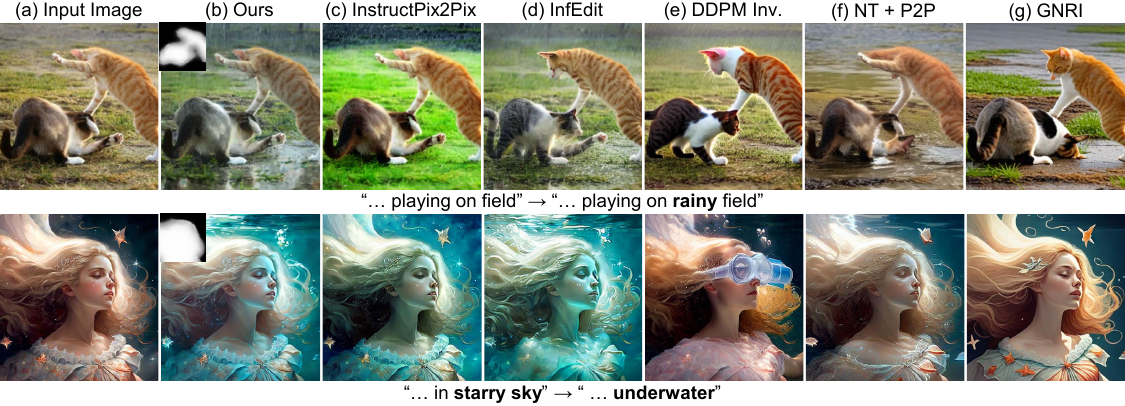}
\end{center}
\vspace{-3mm}
\caption{
\textbf{Qualitative comparison of local editing tasks.}
Our method can generate an edit mask from the text prompt (b, top-left) to enable precise local editing. Other methods (c-g) fail to preserve the structure of the content shared between the source and target prompts.
}
\vspace{-2.0mm}
\label{fig:qual_local_structure_change}
\end{figure*}

%% file: tables/diffusion_editing_quan.tex
\begin{table}[t!]
    \centering
    \resizebox{\columnwidth}{!}{
        \begin{tabular}{lccccccc}
            \toprule
            \multirow{2}{*}{\textbf{Method}} & \multicolumn{5}{c}{\textbf{Preservation}} & \multicolumn{2}{c}{\textbf{Prompt fidelity}} \\ 
            \cmidrule(lr){2-6} \cmidrule(lr){7-8}
            & \textbf{SPL ($\times 10^2$)$\downarrow$} & \textbf{SSIM$\uparrow$} & \textbf{LPIPS$\downarrow$} & \textbf{FSIM$\uparrow$} & \textbf{GMSD$\downarrow$} & \textbf{CLIP S.$\uparrow$} & \textbf{CLIP D.$\uparrow$} \\
            \midrule
            Ours & \textbf{0.107} & \textbf{0.854} & \textbf{0.212} & \textbf{0.927} & \textbf{0.062} & \underline{0.272} & \textbf{0.163} \\
            InfEdit (Baseline) ~\cite{InfEdit_Xu_2024_CVPR} & 0.748 & 0.600 & 0.375 & 0.792 & 0.180 & \textbf{0.284} & \textbf{0.163} \\
            IP2P~\cite{InstructPix2Pix_Brooks_2023_CVPR} & 0.517 & 0.668 & 0.348 & 0.839 & 0.134 & 0.262 & 0.153 \\
            DDPMInv ~\cite{DDPMInversion_Huberman-Spiegelglas_2024_CVPR} & 0.710 & 0.702 & 0.238 & 0.841 & 0.161 & 0.271 & 0.091 \\
            NT+P2P ~\cite{Null-text_Mokady_2023_CVPR, P2P_hertz2023prompttoprompt} & \underline{0.545} & \underline{0.736} & \underline{0.214} & \underline{0.881} & \underline{0.126} & 0.256 & 0.135 \\
            GNRI ~\cite{samuel2025lightningfast_GNRI} & 0.972 & 0.635 & 0.286 & 0.801 & 0.184 & 0.247 & 0.066 \\
            \bottomrule
        \end{tabular}
    }
    \caption{
    \textbf{Quantitative comparison of LDM-based image editing on PIE-Bench} Our method outperforms other models in LDM-based image editing, scoring the highest in preservation metrics. It also stays faithful to the edit prompt, as shown in the prompt fidelity scores.
    }
    \label{tab:diffusion_quan}
\end{table}

%% file: tables/supple_anyedit.tex
\begin{table}[t!]
    \centering
    \resizebox{\columnwidth}{!}{
        \begin{tabular}{lccccccc}
            \toprule
            \multirow{2}{*}{\textbf{Method}} & \multicolumn{5}{c}{\textbf{Preservation}} & \multicolumn{2}{c}{\textbf{Prompt fidelity}} \\ 
            \cmidrule(lr){2-6} \cmidrule(lr){7-8}
            & \textbf{SPL$\downarrow$} & \textbf{SSIM$\uparrow$} & \textbf{LPIPS$\downarrow$} & \textbf{FSIM$\uparrow$} & \textbf{GMSD$\downarrow$} & \textbf{CLIP S.$\uparrow$} & \textbf{CLIP D.$\uparrow$} \\
            \midrule
            Ours & \textbf{0.085} & \textbf{0.906} & \textbf{0.120} & \textbf{0.953} & \textbf{0.046} & 0.263 & 0.078 \\
            InfEdit (Baseline) ~\cite{InfEdit_Xu_2024_CVPR} & 0.865 & 0.648 & 0.280 & 0.813 & 0.176 & \textbf{0.277} & \underline{0.103} \\
            IP2P~\cite{InstructPix2Pix_Brooks_2023_CVPR} & \underline{0.576} & 0.683 & 0.307 & 0.840 & 0.140 & 0.261 & \textbf{0.118} \\
            DDPMInv ~\cite{DDPMInversion_Huberman-Spiegelglas_2024_CVPR} & 0.838 & 0.711 & 0.217 & 0.838 & 0.168 & \underline{0.276} & 0.077 \\
            NT+P2P ~\cite{Null-text_Mokady_2023_CVPR, P2P_hertz2023prompttoprompt} & 0.637 & \underline{0.759} & \underline{0.170} & \underline{0.887} & \underline{0.125} & 0.265 & 0.084 \\
            GNRI ~\cite{samuel2025lightningfast_GNRI} & 1.120 & 0.669 & 0.229 & 0.816 & 0.179 & 0.253 & 0.050 \\
            \bottomrule
        \end{tabular}
    }
    \caption{
    \textbf{Quantitative comparison of LDM-based image editing on AnyEdit benchmark.}
    Consistent with our findings on PIE-Bench, our method demonstrates state-of-the-art performance in all preservation metrics while maintaining competitive prompt fidelity.
    }
    \label{tab:anyedit_quan}
\end{table}

%% file: figures/ablation_loss_fig.tex
\begin{figure}[t!]
\begin{center}
\includegraphics[width=\linewidth]{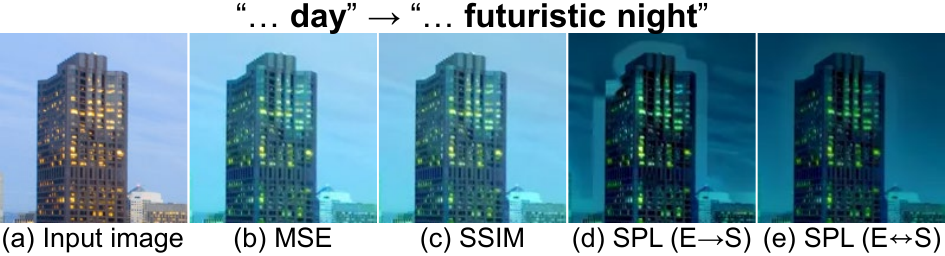}
\end{center}
\vspace{-3mm}
\caption{
\textbf{Ablation study on different loss functions for optimization.}
While other structural losses (b-d) penalizes valid appearance changes (e.g., brightness and contrast), our proposed loss (d) successfully disentangles structure from appearance, enabling a faithful edit while preserving structural fidelity.
}
\label{fig:ablation_loss}
\end{figure}

%% file: figures/ablation_SPL_component.tex
\begin{figure}[t!]
\begin{center}
\includegraphics[width=\linewidth]{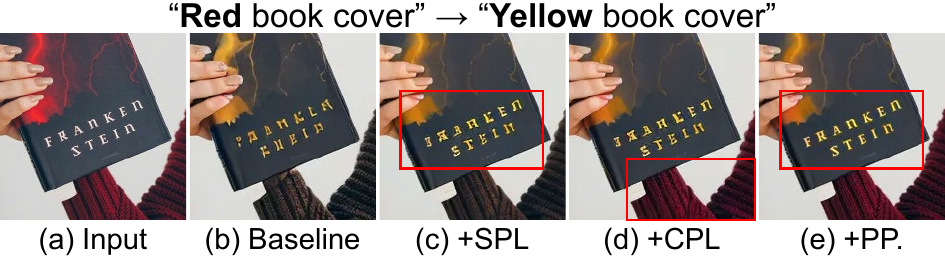}
\end{center}
\vspace{-3mm}
\caption{
\textbf{Qualitative component-wise ablation of our editing method.} Each component of our editing method progressively improves the quality of the baseline edit (b) to our final result (e). PP: Post-processing.
}
\vspace{-4.0mm}
\label{fig:ablation_SPL}
\end{figure}

%% file: tables/ablation_component.tex
\begin{table}[t!]
    \centering
    \resizebox{1.0\linewidth}{!}{
        \begin{tabular}{lccccccc}
            \toprule
            \multirow{2}{*}{\textbf{Method}} & \multicolumn{5}{c}{\textbf{Preservation}} & \multicolumn{2}{c}{\textbf{Prompt fidelity}} \\ 
            \cmidrule(lr){2-6} \cmidrule(lr){7-8}
            & \textbf{SPL ($\times 10^2$) $\downarrow$} & \textbf{SSIM$\uparrow$} & \textbf{LPIPS$\downarrow$} & \textbf{FSIM$\uparrow$} & \textbf{GMSD$\downarrow$} & \textbf{CLIP S.$\uparrow$} & \textbf{CLIP D.$\uparrow$} \\
            \midrule
            Baseline & 0.594 & 0.653 & 0.312 & 0.834 & 0.151 & \textbf{0.281} & \textbf{0.172} \\
            +SPL & 0.351 & 0.720 & 0.265 & 0.890 & 0.102 & 0.274 & 0.168 \\
            +CPL  & 0.351 & 0.720 & 0.252 & 0.892 & 0.102 & 0.272 & 0.161 \\
            +PP & \textbf{0.107} & \textbf{0.853} & \textbf{0.209} & \textbf{0.927} & \textbf{0.063} & 0.272 & 0.163 \\
            \bottomrule
        \end{tabular}
    }
    \caption{
        \textbf{Quantitative component-wise ablation of our editing method.} Adding each component progressively improves all preservation scores while maintaining strong prompt fidelity. PP: Post-processing.
    }
    \label{tab:ablation_component}
\end{table}

%% file: figures/ablation_only_pp.tex
\begin{figure}[t!]
\begin{center}
\includegraphics[width=\linewidth]{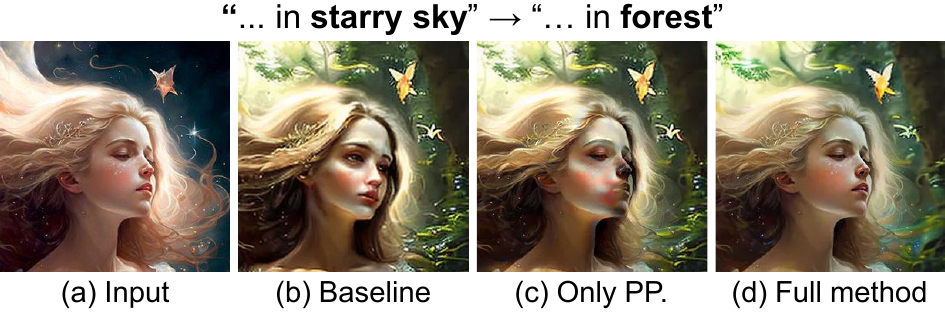}
\end{center}
\vspace{-3mm}
\caption{
\textbf{Guidance vs. Post-Processimg.} Applying our loss only as post-processing (c) fails to fix the baseline's severe structural errors (b). In contrast, our iterative guidance (d) prevents these errors from forming in the first place. PP: Post-processing.
}
\label{fig:ablation_PP}
\end{figure}

%% file: tables/computational_cost.tex
\begin{table}[t!]
    \centering
    \scalebox{0.8}{
        \begin{tabular}{lcc}
            \toprule
            Method & Time(s) & Peak memory(GB) \\ 
            \midrule
            GNRI ~\cite{samuel2025lightningfast_GNRI} & \textbf{0.39} & 13.99 \\
            InfEdit (Baseline) ~\cite{InfEdit_Xu_2024_CVPR} & 2.26 & 5.82 \\
            Ours without PP. & 4.18 & 5.82 \\
            Ours with PP. & 4.58 & 5.82 \\
            IP2P~\cite{InstructPix2Pix_Brooks_2023_CVPR} & 5.29 & \textbf{2.61} \\
            DDPMInv ~\cite{DDPMInversion_Huberman-Spiegelglas_2024_CVPR} & 23.64 & 6.09 \\
            NT+P2P ~\cite{Null-text_Mokady_2023_CVPR, P2P_hertz2023prompttoprompt} & 130.99 & 14.9 \\
            \bottomrule
        \end{tabular}
    }
    \caption{
    \textbf{Computational cost comparison.} Our method demonstrates competitive performance, being substantially faster than inversion-based methods like DDPMInv and NT+P2P, while adding a modest overhead to the fast InfEdit baseline for significantly improved structure preservation. PP: Post-processing.
    }
    \vspace{-4.0mm}
    \label{tab:computational}
\end{table}

%% file: sec/5_conclusion.tex
\section{Conclusion}
We presented a novel structure preservation loss that penalizes the genuine structural differences between two images. Our method integrates this loss into a training-free LDM-based editing pipeline, achieving edits that preserve pixel-level structure. Our approach also includes a color preservation loss and a text-driven edit mask generation scheme for precise local control. Together, these contributions enable a universal image editing method that preserves pixel-level edge structures of the input image, outperforming state-of-the-art baselines as demonstrated in our evaluations.

\paragraph{Limitations and Future Work}
Our method's performance depends on the underlying diffusion model's quality and speed. While our approach will benefit from future advancements in LDMs, its effectiveness may be limited when paired with older or lower-quality backbones. Additionally, the optimization-based application of our loss introduces a computational overhead. We provide an analysis of this overhead and our method's failure cases, including the impact of imperfect masks and challenging edit scenarios, in ~\cref{supple:additional_experiments}. Future work would explore integrating our loss directly into the pre-training of diffusion models to mitigate the computational overhead.

{\small
\paragraph{Acknowledgement}
This research was supported by IITP grants funded by the Korea government (MSIT) (RS-2024-00437866, ITRC(Information Technology Research Center) support program; RS-2019-II191906, Artificial Intelligence Graduate School Program(POSTECH); RS-2024-00395401, Development of VFX creation and combination using generative AI; RS-2024-00457882, AI Research Hub Project), and by the National Research Foundation of Korea (NRF) grant funded by the Korea government (MSIT) (No. 2023R1A2C200494611). Additional support was provided by the Samsung Research Funding \& Incubation Center of Samsung Electronics under Project Number SRFC-IT1801-52.
}

%% file: sec/X_suppl.tex
\clearpage
\appendix
\setcounter{table}{0}
\setcounter{figure}{0}
\setcounter{page}{1}
\maketitlesupplementary

\definecolor{customblue}{rgb}{0.25, 0.41, 0.88} %
{
\setcounter{tocdepth}{2}   %
\hypersetup{linkcolor=customblue}
\tableofcontents
}

\section{Additional Experiments and Analyses}
\label{supple:additional_experiments}

\subsection{Comparison with Task-Specific Structure-Preserving Methods}

In addition to LDM-based editing models, we compare our method against task-specific methods across several key structure-preserving tasks. 

\paragraph{Photo-realistic Style Transfer} synthesizes images by merging content and style from separate images.
We compare our method with PCAKD~\cite{PCA-Based_Knowledge_Distillation_Chiu_2022_CVPR}, utilizing 60 content-style image pairs from the DPST dataset~\cite{Deep_Photo_Style_Transfer_Luan_2017_CVPR}. The evaluation is performed using prompts generated via the method described in~\cref{sec:gpt_prompt_generator}.
In ~\cref{fig:qual_style_transfer}, our method better captures subtle stylistic features compared to PCAKD~\cite{PCA-Based_Knowledge_Distillation_Chiu_2022_CVPR}, producing higher-quality results.

\paragraph{Image Harmonization} adjusts a foreground object’s color and brightness to match a composite image's background. 
Similar to style transfer, we derive prompts using GPT-4o~\cite{gpt4}.
In ~\cref{fig:qual_harmonization}, our method yields results that blend more consistently with the background than PCTNet~\cite{PCT-Net_Guerreiro_2023_CVPR}.

\paragraph{Image Tone Adjustment} modifies the brightness, contrast, and color balance of the input image.
We compare our method with CLIPtone~\cite{CLIPtone_lee2024cliptone} using a subset of~\cite{bychkovsky2011learning}, which is a test set consisting of approximately 500 images and around 50 different tone descriptions. The source and edit text prompt pairs are constructed in the format "a normal photo of..." $\rightarrow$ "a [tone] photo of..." by combining image captions generated by BLIP~\cite{li2022blip} and tone descriptors.
Compared to CLIPtone~\cite{CLIPtone_lee2024cliptone}, our method more accurately and naturally achieves the intended tone.

\paragraph{Seasonal change} alter environmental contexts.
We compare with CycleGAN’s~\cite{CycleGAN} pre-trained summer-to-winter model, using approximately 550 provided test set images. The source and edit text prompt pairs follow the format "a photo of summer ..." $\leftrightarrow$ "a photo of winter ...".
~\cref{fig:qual_season_change} demonstrate our method maintains pixel-level structures and achieves better edit quality compared to CycleGAN, thanks to our diffusion-based prior.

\paragraph{Time-lapse Editing} alter temporal contexts. 
We evaluate against Pix2pix~\cite{pix2pix_Isola_2017_CVPR}, using its pre-trained day-to-night model. We use the night-to-day dataset of~\cite{Transient_Attributes_10.1145/2601097.2601101}, and use 350 daytime images. The source and edit text prompt pairs are manually created in the form "a photo of ... at day" $\rightarrow$ "a photo of ... at night".
~\cref{fig:qual_time_lapse} demonstrate our method maintains pixel-level structures and achieves better edit quality compared to Pix2pix thanks to our diffusion-based prior.

\paragraph{Quantitative Comparison.}
For the evaluation, we use the same metrics used in ~\cref{ssec:ev_det}.
As shown in~\cref{tab:task_specific_quan}, our model achieves superior prompt fidelity compared to methods specifically designed for each structure-preserving image editing task, with notable advantages in structural preservation through our SPL.
LPIPS, being a perceptual metric, may be lower for our model compared to CLIPtone due to our stronger emphasis on structure preservation rather than perceptual similarity alone. 
As discussed in~\cref{subsec:analysis}, we note that SSIM scores can be misleading in tasks involving significant brightness changes (e.g., image tone adjustment or time-lapse), since SSIM strongly penalizes brightness variations.

\subsection{Qualitative Comparison on the AnyEdit Benchmark}
We provide qualitative comparisons on the AnyEdit benchmark~\cite{Yu_2025_CVPR_AnyEdit} in~\cref{fig:supple_qual_global_color_change} and ~\cref{fig:supple_qual_local_structure_change} further support these findings. The visualizations highlight our method's ability to preserve fine-grained, pixel-level edge structures, whereas competing methods often introduce noticeable structural artifacts or fail to fully respect the source image's pixel-level edge structures. Overall, these results strongly corroborate the conclusions drawn from our main experiments on the PIE-Bench subset.

\input{figures/qual_style_transfer}

\input{figures/qual_harmonization}
\input{figures/qual_tone_change}
\input{figures/qual_season_change}
\input{figures/qual_time_lapse}

\input{tables/task_specific_quan}

\input{figures/supple_qual_global_color_change_fig}
\input{figures/supple_qual_local_structure_change}
\input{figures/supple_mask_sdxl}
\input{figures/supple_mask_flux}

\subsection{Cross Attention Mask Upsampling with Different Backbones}
Our cross-attention mask upsampling method extends naturally to various diffusion model architectures beyond the original LDM~\cite{stable_diffusion}, such as SDXL~\cite{sdxl} and FLUX~\footnote{https://huggingface.co/black-forest-labs/FLUX.1-dev}. For U-Net based backbones like SDXL, we follow Prompt-to-Prompt~\cite{P2P_hertz2023prompttoprompt} and extract the cross-attention maps of resolution 32$\times$32  from the bottleneck layer. For FLUX, which has a DiT~\cite{peebles2023scalable_DIT} structure, we extract the average attention map of resolution 64$\times$64 from intermediate blocks 12 through 18. Our guided upsampling algorithm then refines these coarse initial maps to the model's native output resolution—from 16$\times$16 to 512$\times$512 for the original LDM, and 32$\times$32 to 1024$\times$1024 for SDXL and 64$\times$64 to 1024$\times$1024 FLUX. This demonstrates the flexibility of our approach in adapting to different architectures. We show qualitative results for SDXL in~\cref{fig:supple_mask_sdxl} and for FLUX in~\cref{fig:supple_mask_flux}, respectively.

\subsection{Failure Case Analysis}
\paragraph{Cross-Attention Mask Upsampling}
As demonstrated in ~\cref{fig:supple_limitation_mask}, our guided upsampling technique is generally effective at capturing object silhouette with high fidelity. However, the accuracy of the final mask is fundamentally dependent on the quality of the initial coarse cross-attention map. In some cases, if the initial map is not well-localized, the refined mask may cover a slightly larger region than the target object, as shown in the red box of ~\cref{fig:supple_limitation_mask}. When this occurs, SPL is inadvertently applied to this slightly oversized region. While the resulting edit still adheres to the prompt, this can cause subtle edge structures from the source image to be preserved.
\paragraph{Structure Preservation Loss}
The limitation of SPL becomes apparent in tasks that are inherently ambiguous for a structure-preserving method, where an edit may not be entirely aligned with the user's intention. We demonstrate this with a challenging material editing task: transforming a leather bag into a denim one, shown in ~\cref{fig:supple_limitation_SPL}.
As intended, SPL successfully preserves the bag’s overall macro-structure, such as the arm strap and its buckles, and also the fine-grained edge details of the original material's texture. The final edited result is coherent and appears natural with appropriate shading. 
However, if a user intends a structure-breaking material change that completely replaces the micro-texture, such an edit falls outside the intended scope of SPL.

\input{figures/supple_mask_limitation}
\input{figures/supple_SPL_limitation}

\input{figures/distortions_fig}
\input{figures/attention_SPL_scheduling_fig}
\input{tables/similarity_metric}

\input{figures/qual_nt+p2p_spl}
\input{figures/qual_more_examples}

\subsection{
Scheduling of Attention Conditioning and Structure Preservation Loss.}
We explore how the scheduling of attention conditioning and structure preservation loss affects edits.
As we can see in ~\cref{fig:attention_SPL_scheduling}, keeping the attention conditioning strengthens coarse structure preservation, but even with full attention conditioning, the fine structural details of the input image are distorted. However, when combined with structure preservation loss, we can see that the pixel-level edge structural fidelity is kept. 

\subsection{Validation as a Structural Difference Metric}
To validate our proposed loss as a robust metric for structural similarity, we evaluate its response to a range of image distortions as seen in ~\cref{fig:distortions}. The results in ~\cref{tab:similarity_metric} demonstrate that our metric successfully disentangles structure from appearance. It registers a low penalty for non-structural distortions (e.g., color and brightness shifts) while correctly identifying structural distortions. In contrast, common metrics like SSIM~\cite{SSIM} and LPIPS~\cite{LPIPS} often conflate these two, assigning high penalties to non-structural changes.

\subsection{Generalization to Different Backbones}
To demonstrate the generality and modularity of our approach, we apply our structure-preserving editing method to a different baseline: the combination of Null-text inversion~\cite{Null-text_Mokady_2023_CVPR} and Prompt-to-Prompt~\cite{P2P_hertz2023prompttoprompt}. As shown in ~\cref{fig:qual_nt+p2p_spl}, our method enhances the structural fidelity of the baseline's output, confirming its effectiveness across different editing frameworks.

\subsection{Additional Qualitative Comparison}
We provide additional qualitative comparison results with LDM-based editing methods in ~\cref{fig:qual_more_examples}. We can see that our method persistently achieves the best pixel-level edge structure preservation without the loss of prompt fidelity.

\section{Additional Implementation Details}
For all experiments, we use total inference steps of \( T = 15 \).
For the structure preservation loss, defined via a local linear model in Equation~\cref{eq:local_linear_model}, we configure a window size of \( \omega_k = 11 \) and a regularizer \( \epsilon = 10^{-4} \). In the optimization-driven denoising process, we apply stochastic gradient descent with a learning rate \( \eta = 1 \) and momentum 0.9. During optimization we use \( \lambda = 10^{-4} \) for to emphasize structural fidelity (\cref{eq:diffusion_with_SPL2}). We fix the number of optimization iterations at \( k = s = 100 \) and structure preservation loss threshold timestep \( t_{\text{SPL}} = 12 \).
For tasks requiring localized edits, we employ the upscaled masks from ~\cref{subsec:cross_attention_mask}.  All experiments were conducted on an NVIDIA A6000 GPU. The models used in our experiments are as follows: for InfEdit~\cite{InfEdit_Xu_2024_CVPR}, we used LCM Dreamshaper v7; for InstructPix2Pix~\cite{InstructPix2Pix_Brooks_2023_CVPR}, the official model provided by the authors was employed. Both DDPMInv~\cite{DDPMInversion_Huberman-Spiegelglas_2024_CVPR} and NT+P2P~\cite{Null-text_Mokady_2023_CVPR, P2P_hertz2023prompttoprompt} were implemented using Stable Diffusion v1.4.

When applying the coarse structure preservation through attention conditioning as detailed in ~\cref{subsec:structure_preserving_ldm}, we observe that applying this attention conditioning across all denoising timesteps can overly constrain the latent, reducing fidelity to the edit prompt \(p_{\text{edit}}\). To balance coarse structure preservation with edit flexibility, we schedule the attention conditioning, applying \(f^{\text{src}}_t\) only for timesteps \(t \geq t_{\text{attn}}\). We set this attention conditioning scheduling timestep as $t_{\text{attn}}= 12$.

\section{Derivation of the LLM Coefficients}
Given images $I^E$ and $I^S$, the local linear model defines the relationship between these two images within a local window $\omega_k$ as:
\begin{equation}
    I^S_i = a_k \, I^E_i + b_k, \quad \forall i \in \omega_k,
    \label{eq:local_linear_model_supple}
\end{equation}

We derive the coefficients $a_k$ and $b_k$ by minimizing the cost function $E(a_k, b_k)$ within a local window $\omega_k$:
\begin{equation}
    E(a_k, b_k) = \sum_{i \in \omega_k} \left( (a_k I^E_i + b_k) - I^S_i \right)^2.
\end{equation}
The minimum is found by setting the partial derivatives with respect to $a_k$ and $b_k$ to zero.

\paragraph{Derivation of the Offset Coefficient $b_k$}
Differentiating $E$ with respect to $b_k$ and setting the result to zero yields:
\begin{align*}
    &\frac{\partial E}{\partial b_k} = 2 \sum_{i \in \omega_k} (a_k I^E_i + b_k - I^S_i) = 0 \\
    \implies &a_k \sum_{i \in \omega_k} I^E_i + \sum_{i \in \omega_k} b_k - \sum_{i \in \omega_k} I^S_i = 0
\end{align*}
Dividing by $|\omega_k|$, the number of pixels in the window, gives the means $\mu_k^E$ and $\mu_k^S$. Hence, solving for $b_k$:
\begin{align*}
    &a_k \mu^E_k + b_k - \mu^S_k = 0 \\
    \implies &b_k = \mu^S_k - a_k \mu^E_k.
\end{align*}

\paragraph{Derivation of the Scaling Coefficient $a_k$}
Next, we differentiate $E$ with respect to $a_k$ and set the result to zero:
\begin{equation*}
    \frac{\partial E}{\partial a_k} = 2 \sum_{i \in \omega_k} (a_k I^E_i + b_k - I^S_i)I^E_i = 0
\end{equation*}
Substituting our expression for $b_k = \mu^S_k - a_k \mu^E_k$:
\begin{align*}
    &\sum_{i \in \omega_k} (a_k I^E_i + (\mu^S_k - a_k \mu^E_k) - I^S_i)I^E_i = 0 \\
    \implies &\sum_{i \in \omega_k} (a_k(I^E_i - \mu^E_k) - (I^S_i - \mu^S_k))I^E_i = 0
\end{align*}
Solving for $a_k$:
\begin{equation*}
    a_k = \frac{\sum_{i \in \omega_k} (I^S_i - \mu^S_k)I^E_i}{\sum_{i \in \omega_k} (I^E_i - \mu^E_k)I^E_i}
\end{equation*}
This expression is equivalent to the covariance of $I^E$ and $I^S$ divided by the variance of $I^E$. Dividing the numerator and denominator by $|\omega_k|$ and including the regularization term $\rho$, we arrive at the final form:
\begin{equation}
    a_k = \frac{\frac{1}{|\omega_k|} \sum_{i \in \omega_k} I^E_i I^S_i - \mu^E_k \mu^S_k}{(\sigma^E_k)^2 + \rho},
\end{equation}
where $(\sigma^E_k)^2$ is the variance of $I^E$ in $\omega_k$.

\section{Algorithms}
\begin{algorithm*}
\caption{Optimization-Driven Denoising Process}
\label{alg:guided_upsample}
\begin{algorithmic}[1]
\Require Source image $I_{\text{src}}$, edit prompt $p_{\text{edit}}$, source features \(f^{\text{src}}_t\), noise prediction model $\epsilon_\theta$, encoder $\mathcal{E}$, decoder $\mathcal{D}$, maximum timestep $T$, coefficients $a_t, b_t$ from noise schedule, scheduling timestep threshold $t_s$, learning rates $\eta, w$, number of optimization steps $k, s$, loss weights $\lambda$
\Ensure Edited image $\hat{I}_0$

\State Initialize latent $z_T \sim \mathcal{N}(0, \mathbf{I})$

\For{$t$ in $(T, 1)$}
    \State $\hat{\epsilon}_t \gets \epsilon_\theta\left(z_t, t, p_{\text{edit}}, f^{\text{src}}_t\right)$
    \State $\hat{z}_0^{(t)} \gets \frac{1}{a_t} \left( z_t - b_t \hat{\epsilon}_t \right)$
    \If{$t \leq t_s$}       \Comment{Scheduling optimization to preserve details}
        \State $\hat{I} \gets \mathcal{D}(\hat{z}_0^{(t)})$ \Comment{Decode to image space}
        \For{$i$ in $(0, k)$}       \Comment{Gradient descent step}
            \State $\hat{I} \leftarrow \hat{I} - \eta \nabla_{\hat{I}}\left\{ \mathcal{L}_{\text{SPL}}(I_{\text{src}}, \hat{I}) + \lambda\mathcal{L}_{\text{CPL}}(I_{\text{src}}, \hat{I})\right\}$
        \EndFor
        \State $\tilde{z}_0^{(t)} \gets \mathcal{E}(\hat{I})$ \Comment{Re-encode optimized image}
    \EndIf
    \State $z_{t-1} \gets \mathcal{S}\left(\tilde{z}_0^{(t)}, z_t, t, \hat{\epsilon}_t \right)$
\EndFor
\State $\hat{I}_0 \gets \mathcal{D}(z_0)$
\For{$i$ in $(0, s)$} \Comment{Post processing in image space}
    \State $\hat{I}_0 \gets \hat{I}_0 - \eta \nabla_{\hat{I}_0}\left\{ \mathcal{L}_{\text{SPL}}(I_{\text{src}}, \hat{I}_0) + \lambda\mathcal{L}_{\text{CPL}}(I_{\text{src}}, \hat{I}_0)\right\}$
\EndFor
\end{algorithmic}
\end{algorithm*}

\begin{algorithm*}[htbp]
\caption{Iterative Guided Mask Upsampling}
\label{alg:guided_upsample}
\begin{algorithmic}[1]
\Require Initial cross-attention map $M_{init}$, reference image $I$, target size $T$, initial radius $r$, radius increment $\Delta r$
\Ensure Refined mask $M$

\State $M \gets \text{Binarize}(M_{init}, 0.4)$
\State $s \gets \text{size}(M)$ \Comment{Get initial resolution.}

\While {$s < T$}
    \State $s \gets 2 \times s$
    \State $M \gets \text{BilinearUpsample}(M, \text{scale}=2)$
    \State $I_s \gets \text{Resize}(I, s)$ \Comment{Downscale reference image to $s \times s$ resolution.}
    \State $M \gets \text{GuidedFilter}(M, I_s, r, \epsilon)$
    \State $r \gets r + \Delta r$
\EndWhile

\State \textbf{return} $M$

\end{algorithmic}
\end{algorithm*}

We provide the overall algorithm of the optimization-driven denoising process in ~\cref{subsec:structure_preserving_ldm} in ~\cref{alg:guided_upsample}.
We provide the cross-attention mask upsampling algorithm of ~\cref{subsec:cross_attention_mask} in ~\cref{alg:guided_upsample}.

\section{Source and Edit Prompt Generation for Image-Based Editing Tasks}
\label{sec:gpt_prompt_generator}
\input{figures/gpt_prompt_image_harmonization}
\input{figures/gpt_prompt_style_transfer}

Unlike other image editing tasks where a text prompt is provided or can be easily specified, image harmonization and photorealistic style transfer depend on an additional input image that visually encodes the editing instructions. This creates a challenge for text-based editing models, which require these visual instructions to be converted into text prompts. To address this, we employ a multi-modal large language model, such as GPT-4o~\cite{gpt4}, drawing inspiration from the prompt generation approach in Diff-Harmonization~\cite{chen2025zeroshot}.

\paragraph{Image Harmonization.}
In image harmonization, our aim is to seamlessly integrate a foreground object into a background. We begin by using the mask image to distinguish the foreground and background regions. Next, a vision-language model generates text descriptions for the foreground object (FO), its foreground description (FD), and the background description (BD). Using these, we construct the source prompt (``FD FO'') and edit prompt (``BD FO'').
The specific template for this is illustrated in~\cref{fig:gpt_prompt_image_harmonization}, where we feed this prompt into GPT-4 to produce the source and edit prompts. 

\paragraph{Photorealistic Style Transfer.}
In photorealistic style transfer, the goal is to apply the visual style of one image to the content of another. 
We start by using a vision-language model to create text descriptions for both the content image and the style image. 
From the style image’s description, we extract terms that capture its visual style. 
Then, we modify the content image’s description by replacing its style-related terms with those derived from the style image.
Following the template in~\cref{fig:gpt_prompt_style_transfer}, GPT-4o generates a source and edit text prompt pair based on this approach.

\section{Detailed Related Works on Task-Specific Structure-Preserving Image Editing}
We provide additional details on long-standing image editing tasks where it is crucial to preserve the pixel-level structure of the input image. 
We also briefly discuss the approaches for these tasks that were introduced before diffusion-based image editing. 
While the results produced by these earlier methods show high structural fidelity due to the specific assumptions they make, this specialization restricts their use in broader editing scenarios.

\textit{Image Relighting} modifies the illumination of an input image. 
Recent learning-based approaches primarily rely on physics-based priors and image datasets obtained from a light-stage~\cite{Acquiring_the_reflectance_field_of_a_human_face_10.1145/344779.344855, Performance_relighting_and_reflectance_wenger2005performance} to achieve realistic relighting results while preserving the underlying scene ~\cite{Deep_Single-Image_Portrait_Relighting_Zhou_2019_ICCV, Total_Relighting_10.1145/3450626.3459872, SwitchLight_kim2024switchlight, sun2019single, nestmeyer2020learning}. Despite their effectiveness, these methods remain specialized for physics-based relighting tasks.

\textit{Image Tone Adjustment} alters tonal properties—such as brightness, contrast, and color balance—of the input image while preserving its structure. 
Techniques range from color transformation matrix-based methods~\cite{chai2020supervised, gharbi2017deep, yan2016automatic} to look-up-table-based methods~\cite{he2020conditional,kim2021representative,wang2021real,zeng2020learning, CLIPtone_lee2024cliptone}. These methods effectively constrain edits to color space transformations and thereby preserve edges and spatial structure. 
However, their reliance on fixed transformations limits adaptability to more complex tasks.

\textit{Image Harmonization} and \textit{Background Replacement} aim to make a composite image visually coherent by adjusting the foreground to match the color statistics and illumination of the background. 
Traditionally, image gradient-based methods~\cite{jia2006drag, 10.1145/1201775.882269, sunkavalli2010multi, tao2013error} and image color statistics-based methods~\cite{10.1145/1179352.1141933, pitie2005n, reinhard2001color, xue2012understanding} were proposed, followed by data-driven methods using neural networks~\cite{ling2021region, PCA-Based_Knowledge_Distillation_Chiu_2022_CVPR}.
While these methods mostly preserve the input images' structure, they focus narrowly on compositing scenarios.

\textit{Photorealistic Style Transfer} transfers the reference style onto the input image while preserving style-independent features of the input image. 
Early works operated similarly to image harmonization by matching the image statistics of the input and reference images~\cite{reinhard2001color, pitie2005n}.
Modern approaches match the statistics of deep features~\cite{gatys2016image, li2016combining}, where these features are obtained by feeding the image into a pretrained image classification neural network~\cite{simonyan2014very}.
Follow-up works have addressed the structural artifacts produced during style transfer, aiming to retain fine structural details of the input image~\cite{Deep_Photo_Style_Transfer_Luan_2017_CVPR,yoo2019photorealistic, li2018closed, PCA-Based_Knowledge_Distillation_Chiu_2022_CVPR}.
However, these methods do not account for other types of attribute manipulation other than the overall image style.

\textit{Time-Lapse} and \textit{Season or Weather Change} involve hallucinating how a scene would appear with different transient attributes, such as time or weather. 
Data-driven algorithms have suggested example-based appearance transfer~\cite{Data-driven_hallucination_10.1145/2508363.2508419, Transient_Attributes_10.1145/2601097.2601101}, but editing is constrained to domain-specific datasets.
Since modifying transient attributes often requires generating new details, GAN-based models have also been introduced~\cite{pix2pix_Isola_2017_CVPR, ToDayGAN, CycleGAN}.
However, they also rely on domain-specific datasets, limiting their application to a particular setting.

%% file: figures/qual_style_transfer.tex
\begin{figure}[t!] 
\begin{center}
\includegraphics[width=\linewidth]{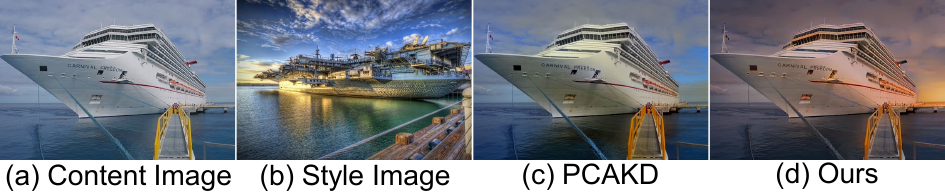}
\end{center}
\caption{
\textbf{Qualitative comparison on photorealistic style transfer}.
Given a content image (a) and a style image (b), our method effectively transfers the sunrise style, producing a naturally stylized result (d). In contrast, PCAKD (c) fails to transfer this style effectively.
}
\label{fig:qual_style_transfer}
\end{figure}

%% file: figures/qual_harmonization.tex
\begin{figure}[t!] 
\begin{center}
\includegraphics[width=\linewidth]{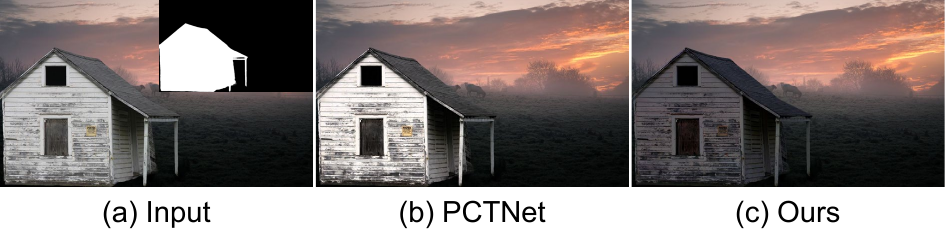}
\end{center}
\caption{
\textbf{Qualitative comparison on image harmonization}.
(a) Composed image with foreground mask (top-right). 
The Result from PCTNet (b) exhibits clear lighting inconsistencies. Our method (c) seamlessly blends the foreground and background.
}
\label{fig:qual_harmonization}
\end{figure}

%% file: figures/qual_tone_change.tex
\begin{figure}[t!]
\begin{center}
\includegraphics[width=\linewidth]{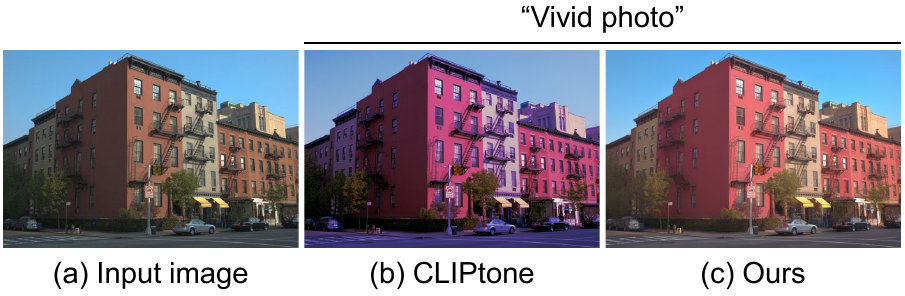}
\end{center}
\caption{
\textbf{Qualitative comparison on tone adjustment}.
While the result of CLIPtone (b) is overly saturated, our method (c) produces a naturally-toned result well-aligned with the text description.
}
\label{fig:qual_tone_change}
\end{figure}

%% file: figures/qual_season_change.tex
\begin{figure}[t!]
\begin{center}
\includegraphics[width=\linewidth]{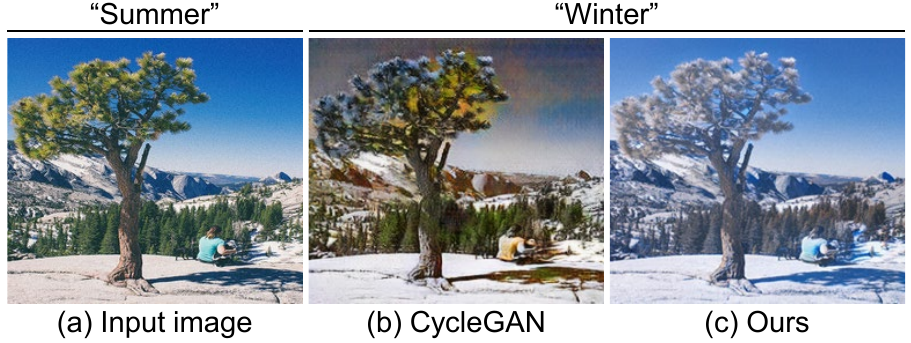}
\end{center}
\caption{
\textbf{Qualitative comparison on season change}. 
While CycleGAN (b) produces an incoherent result, our method (c) successfully generates a natural seasonal transformation, realistically integrating effects like snow.
}
\label{fig:qual_season_change}
\end{figure}

%% file: figures/qual_time_lapse.tex
\begin{figure}[t!]
\begin{center}
\includegraphics[width=\linewidth]{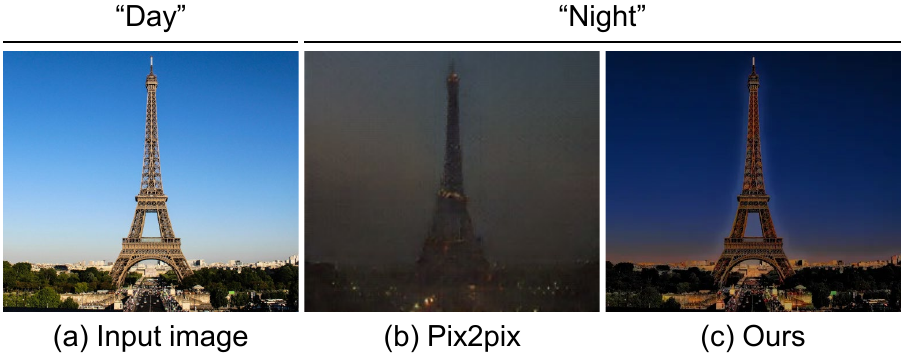}
\end{center}
\caption{
\textbf{Qualitative comparison on time-lapse}.
Pix2Pix (b) introduces significant structural distortions. Our method (c) achieves a realistic time-of-day change while maintaining the structure.
}
\label{fig:qual_time_lapse}
\end{figure}

%% file: tables/task_specific_quan.tex
\begin{table}[t!]
    \centering
    \resizebox{\columnwidth}{!}{
        \begin{tabular}{lccccc}
            \toprule
            \multirow{2}{*}{\textbf{Method}} & \multicolumn{3}{c}{\textbf{Preservation}} & \multicolumn{2}{c}{\textbf{Prompt fidelity}} \\ 
            \cmidrule(lr){2-4} \cmidrule(lr){5-6}
            & \textbf{SPL($\times 10^2$) $\downarrow$} & \textbf{SSIM$\uparrow$} & \textbf{LPIPS$\downarrow$} & \textbf{CLIP S.$\uparrow$} & \textbf{CLIP D.$\uparrow$} \\
            \midrule \midrule
            \multicolumn{6}{c}{Photorealistic Style Transfer} \\
            \midrule
            Ours & \textbf{0.006} & \textbf{0.879} & \textbf{0.182} & \textbf{0.254} & \textbf{0.147} \\
            PCAKD & 0.752 & 0.478 & 0.346 & 0.248 & 0.130 \\
            \midrule \midrule
            \multicolumn{6}{c}{Image Tone Adjustment} \\
            \midrule
            Ours & \textbf{0.010} & 0.680 & 0.250 & \textbf{0.237} & \textbf{0.078} \\
            CLIPTone & 0.016 & \textbf{0.862} & \textbf{0.154} & 0.226 & 0.075 \\
            \midrule \midrule
            \multicolumn{6}{c}{Time-lapse} \\
            \midrule
            Ours & \textbf{0.070} & 0.240 & \textbf{0.471} & 0.234 & \textbf{0.192} \\
            Pix2pix & 0.534 & \textbf{0.396} & 0.623 & \textbf{0.235} & 0.136 \\
            \midrule \midrule
            \multicolumn{6}{c}{Season/Weather Change} \\
            \midrule
            Ours & \textbf{0.061} & \textbf{0.856} & \textbf{0.266} & 0.197 & \textbf{0.170} \\
            CycleGAN & 1.358 & 0.463 & 0.350 & \textbf{0.198} & 0.097 \\
            \bottomrule
        \end{tabular}
        }
    \caption{
    \textbf{Quantitative comparison with task-specific editing methods.} Our method consistently achieves the best structure preservation loss across all tasks while maintaining high prompt fidelity. Note that SSIM is sensitive to luminance and contrast variations; thus, for tasks requiring brightness or contrast adjustments (e.g., tone adjustment, time-lapse), SSIM scores may not accurately reflect structural preservation. Additionally, LPIPS primarily captures perceptual similarity rather than structural fidelity.
    }
    \label{tab:task_specific_quan}
\end{table}

%% file: figures/supple_qual_global_color_change_fig.tex
\begin{figure*}[t!] 
\begin{center}
\includegraphics[width=\linewidth]{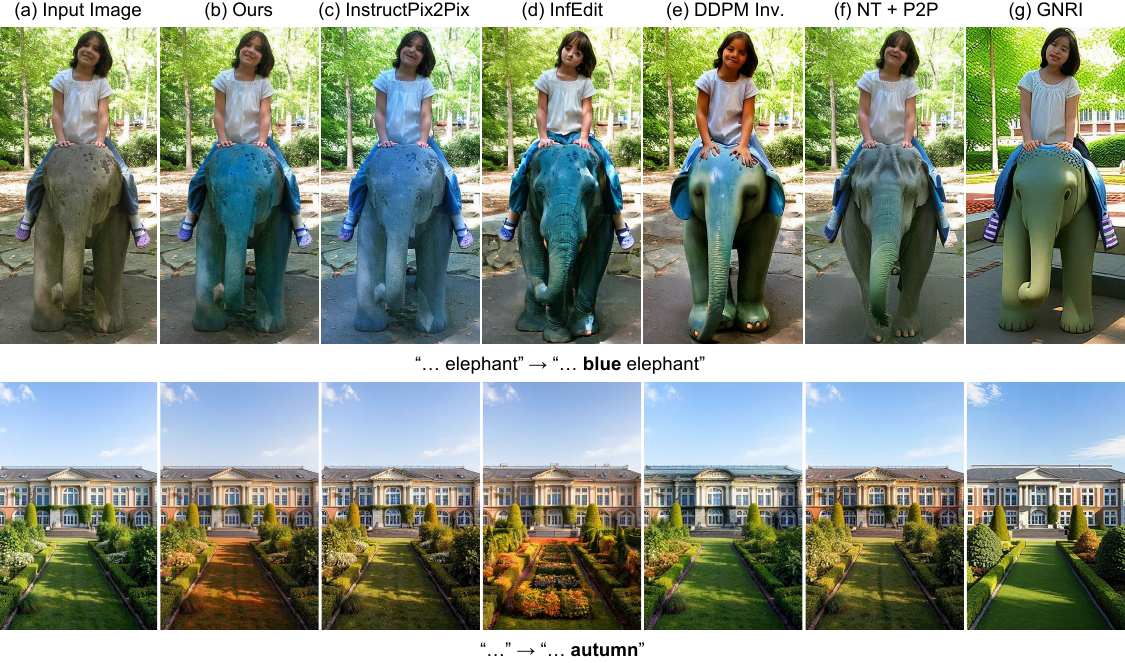}
\end{center}
\vspace{-5.0mm}
\caption{
\textbf{Qualitative comparison on global editing tasks.}
Our method (b) successfully applies the edit while preserving fine-grained structural details. Other methods (c-g) exhibit either low prompt fidelity or significant structural distortions
}
\vspace{-1.0mm}
\label{fig:supple_qual_global_color_change}
\end{figure*}

%% file: figures/supple_qual_local_structure_change.tex
\begin{figure*}[t!] 
\begin{center}
\includegraphics[width=\linewidth]{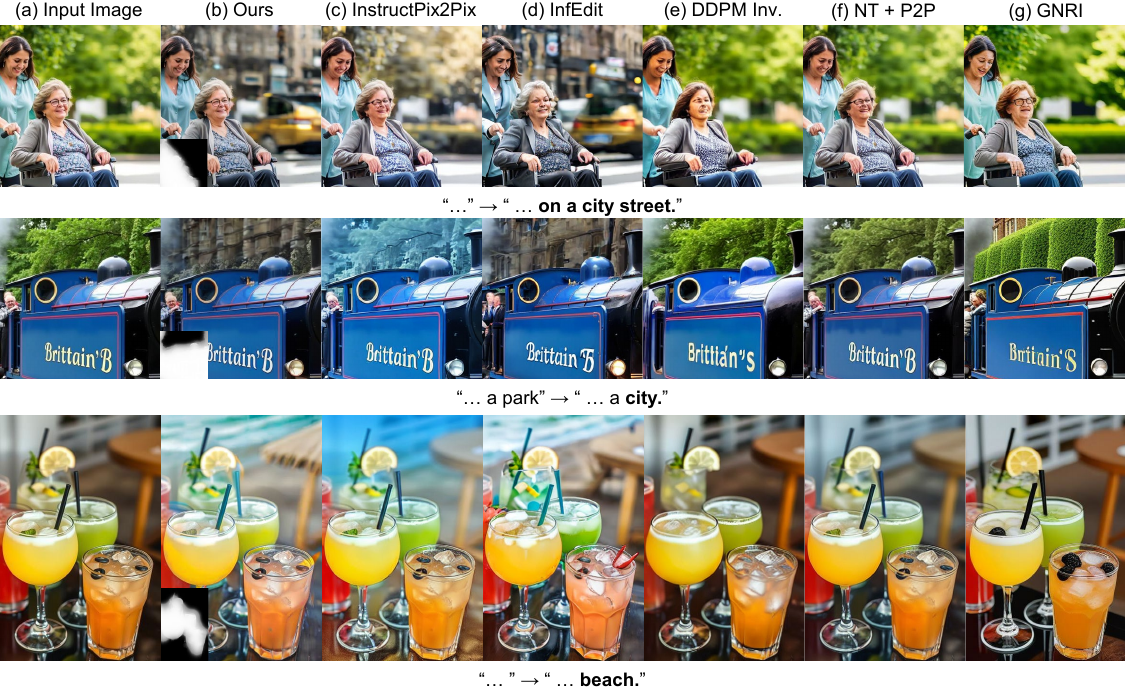}
\end{center}
\vspace{-5.0mm}
\caption{
\textbf{Qualitative comparison of local editing tasks.}
Our method can generate an edit mask from the text prompt (b, bottom-left) to enable precise local editing. Other methods (c-g) fail to preserve the structure of the content shared between the source and target prompts.
}
\label{fig:supple_qual_local_structure_change}
\end{figure*}

%% file: figures/supple_mask_sdxl.tex
\begin{figure}[t!]
\begin{center}
\includegraphics[width=\linewidth]{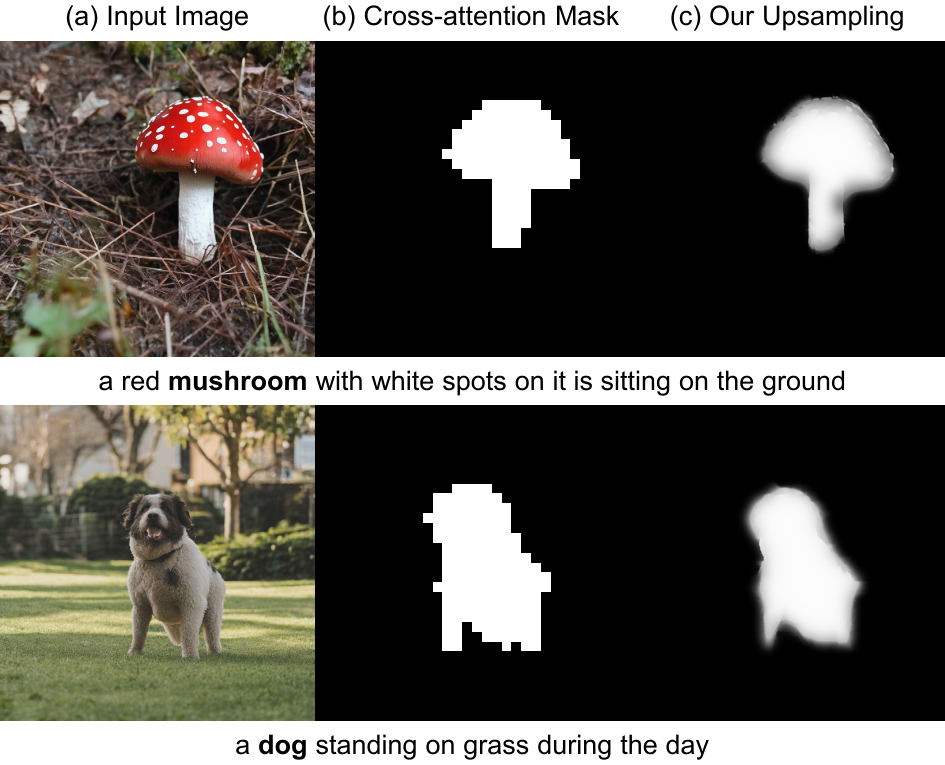}
\end{center}
\vspace{-5.0mm}
\caption{
\textbf{Cross-attention mask upsampling for SDXL~\cite{sdxl}}. By upsampling the coarse attention map (b), our method generates a sharp, high-resolution mask (c).
}
\label{fig:supple_mask_sdxl}
\vspace{-5.0mm}
\end{figure}

%% file: figures/supple_mask_flux.tex
\begin{figure}[t!]
\begin{center}
\includegraphics[width=\linewidth]{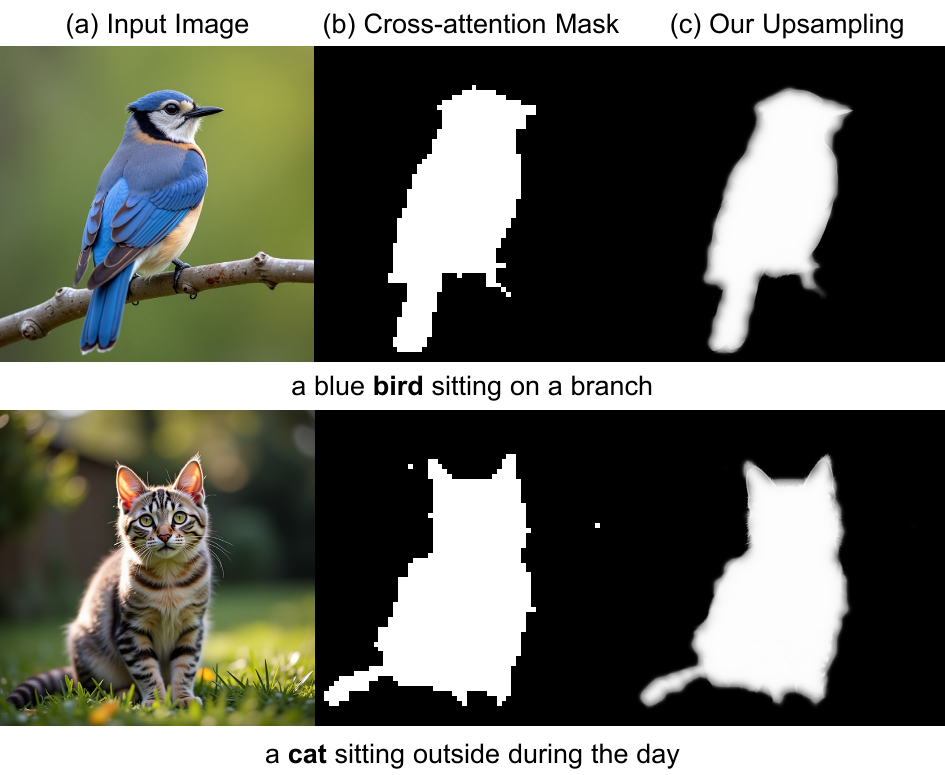}
\end{center}
\vspace{-5.0mm}
\caption{
\textbf{Cross-attention mask upsampling for FLUX.}
By upsampling the coarse attention map (b), our method generates a sharp, high-resolution mask (c).
}
\label{fig:supple_mask_flux}
\vspace{-5.0mm}
\end{figure}

%% file: figures/supple_mask_limitation.tex
\begin{figure}[t!]
\begin{center}
\includegraphics[width=\linewidth]{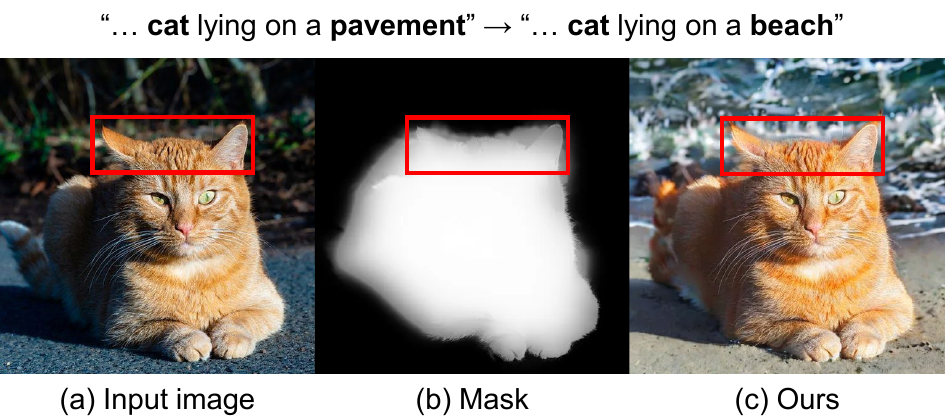}
\end{center}
\caption{
\textbf{Imprecise Edit Mask Example.}
The soft boundaries of the upsampled mask (b) can sometimes extend slightly beyond the foreground object. As a result, subtle structural details from the source image (a) are unintentionally preserved near the cat's silhouette (c).
}
\label{fig:supple_limitation_mask}
\end{figure}

%% file: figures/supple_SPL_limitation.tex
\begin{figure}[t!]
\begin{center}
\includegraphics[width=\linewidth]{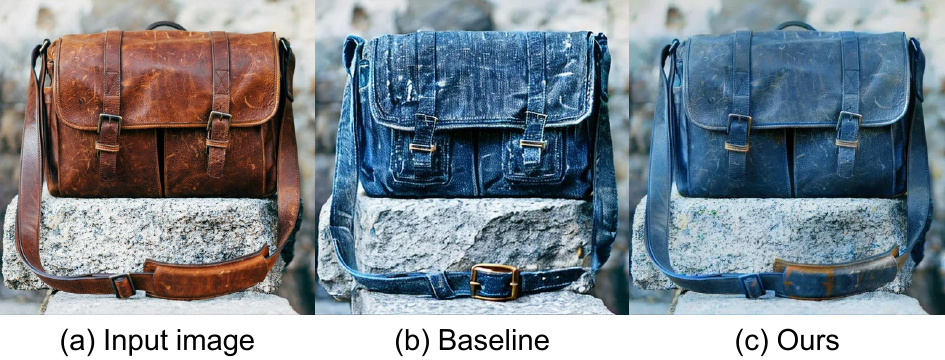}
\end{center}
\caption{
\textbf{Material editing example from leather to denim}. Our method (c) preserves the fine-grained texture and wear patterns of the original leather, while the baseline (b) breaks the structure and replaces the material entirely.
}
\label{fig:supple_limitation_SPL}
\end{figure}

%% file: figures/distortions_fig.tex
\begin{figure}[t!]
\begin{center}
\includegraphics[width=\linewidth]{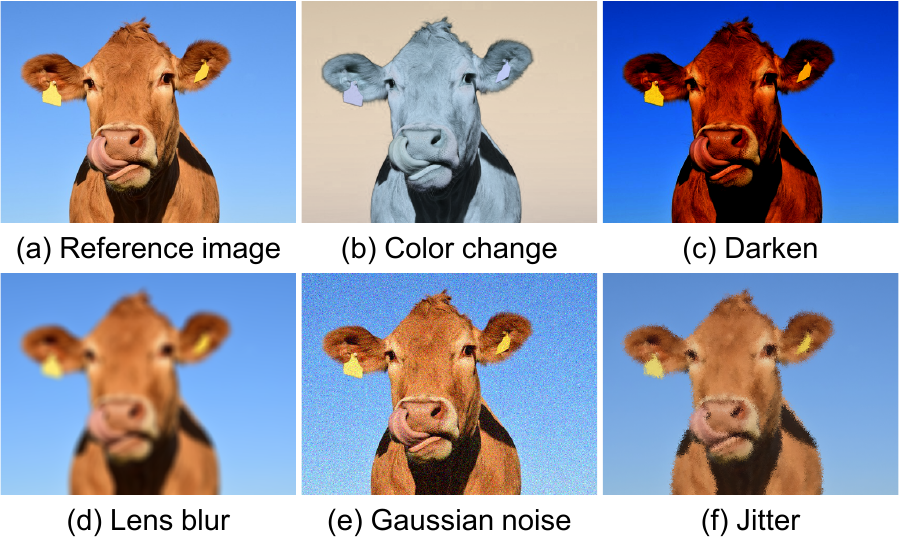}
\end{center}
\vspace{-5.0mm}
\caption{
\textbf{Examples of image distortions.} (b-c) Non-structural distortions, (d-f) Structural distortions.}
\label{fig:distortions}
\vspace{-1.0mm}
\end{figure}

%% file: figures/attention_SPL_scheduling_fig.tex
\begin{figure}[t!]
\begin{center}
\includegraphics[width=\linewidth]{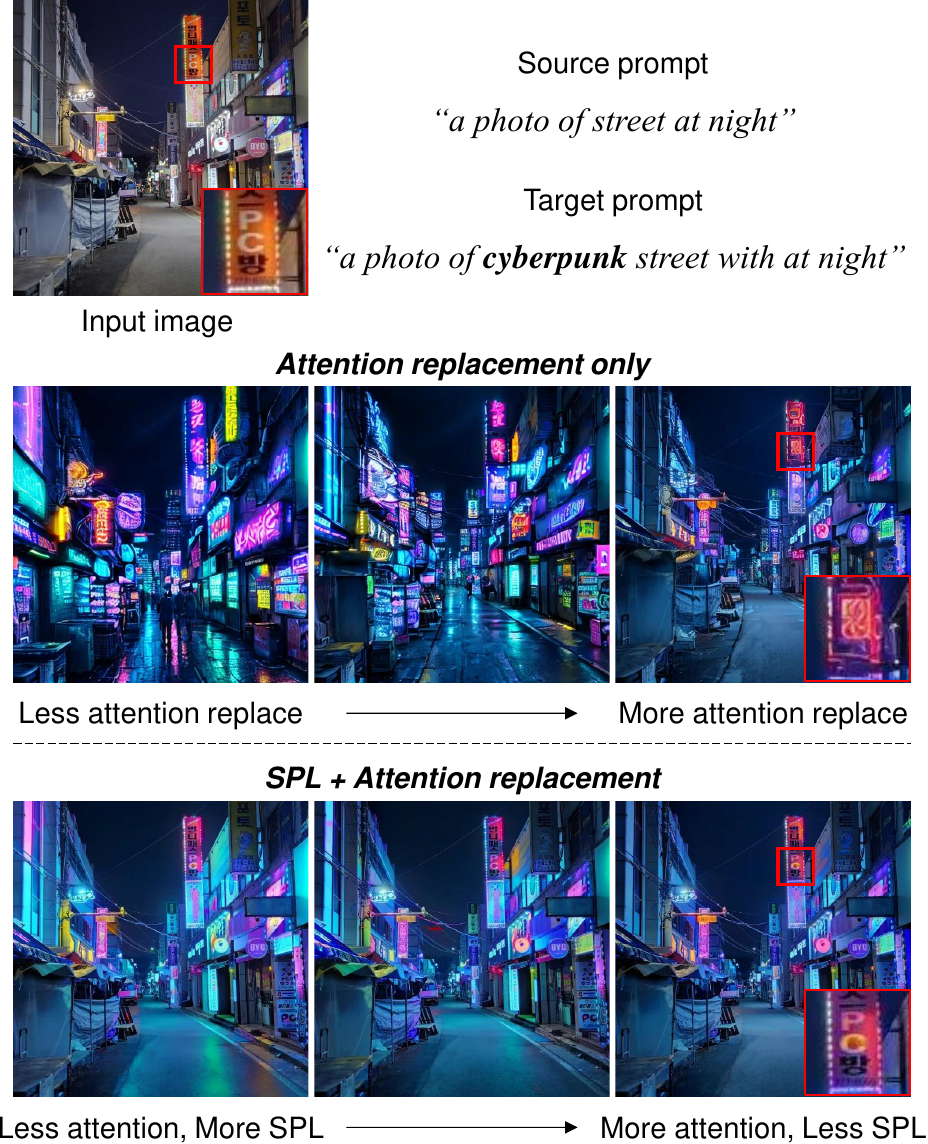}
\end{center}
\caption{
\textbf{Scheduling of attention conditioning and structure preservation loss.} 
We analyze the impact of attention conditioning and the structure preservation loss on structural fidelity. As shown in~\cref{fig:attention_SPL_scheduling}, applying attention conditioning throughout the generative process helps retain coarse structures but fails to preserve fine details, even at full conditioning. In contrast, incorporating our structure preservation loss effectively maintains pixel-level edge fidelity, even with reduced attention conditioning.
}
\label{fig:attention_SPL_scheduling}
\end{figure}

%% file: tables/similarity_metric.tex
\begin{table}[t!]
    \centering
    \resizebox{1.0\linewidth}{!}{
        \begin{tabular}{lcccc}
            \toprule
            \textbf{Distortion} & \textbf{SPL ($\times 10^2$)$\downarrow$} & \textbf{PSNR$\uparrow$} & \textbf{SSIM$\uparrow$} & \textbf{LPIPS$\downarrow$} \\ 
            \midrule
            Color change    & 0.080 & 11.06 & 0.927 & 0.504 \\
            Darken          & 0.063 & 10.60 & 0.664 & 0.195 \\
            Lens blur       & 0.241 & 24.77	& 0.776	& 0.301 \\
            White noise 	& 0.356 & 20.56	& 0.338	& 0.551 \\
            Jitter          & 0.325 & 22.20	& 0.783	& 0.196 \\
            \bottomrule
        \end{tabular}
    }
    \caption{
    \textbf{Comparison of image similarity metrics across different types of distortions.}
    }
    \vspace{-2.0mm}
    \label{tab:similarity_metric}
\end{table}

%% file: figures/qual_nt+p2p_spl.tex
\begin{figure*}[t!] 
\begin{center}
\includegraphics[width=\linewidth]{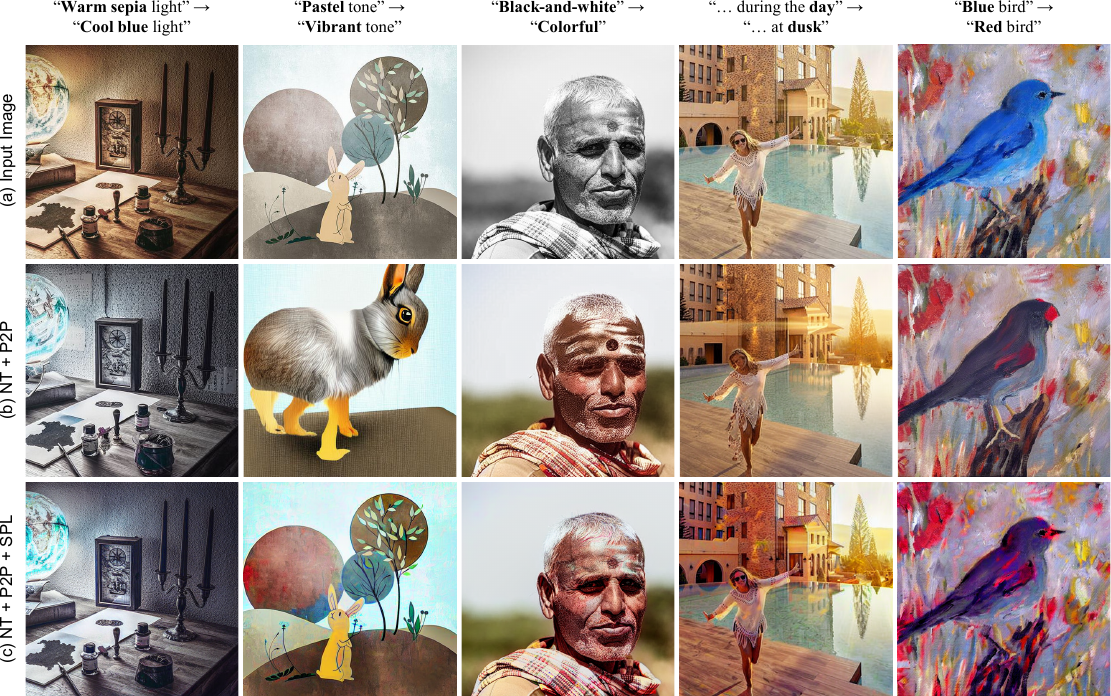}
\end{center}
\caption{
\textbf{Generalizability of our method across different baseline model.} Our method can be integrated into diverse LDM-based image editing pipelines (e.g., Null-text inversion + Prompt-to-Prompt), enhancing their ability to preserve the structural details of the input image during editing.
}
\label{fig:qual_nt+p2p_spl}
\end{figure*}

%% file: figures/qual_more_examples.tex
\begin{figure*}[t!] 
\begin{center}
\includegraphics[width=\linewidth]{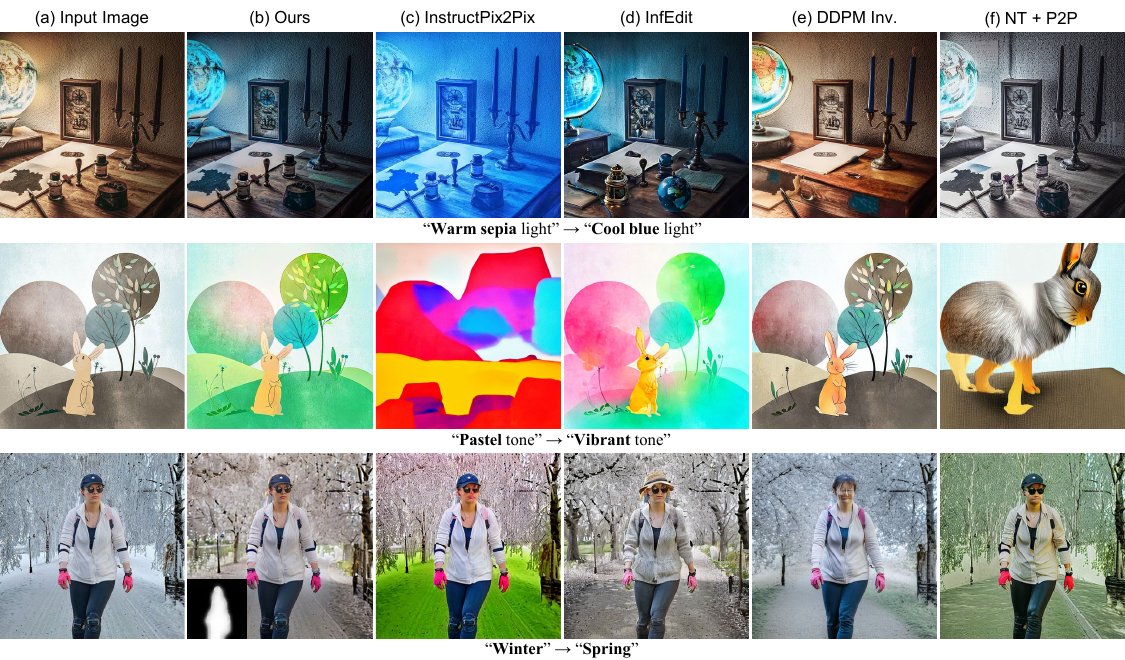}
\end{center}
\caption{
\textbf{Additional qualitative comparison with LDM-based image editing methods}. The first and second rows demonstrate global editing results. The Last row shows additional local editing results.
}
\label{fig:qual_more_examples}
\end{figure*}

%% file: figures/gpt_prompt_image_harmonization.tex
\begin{figure*}[t!] 
\begin{center}
\includegraphics[width=\linewidth]{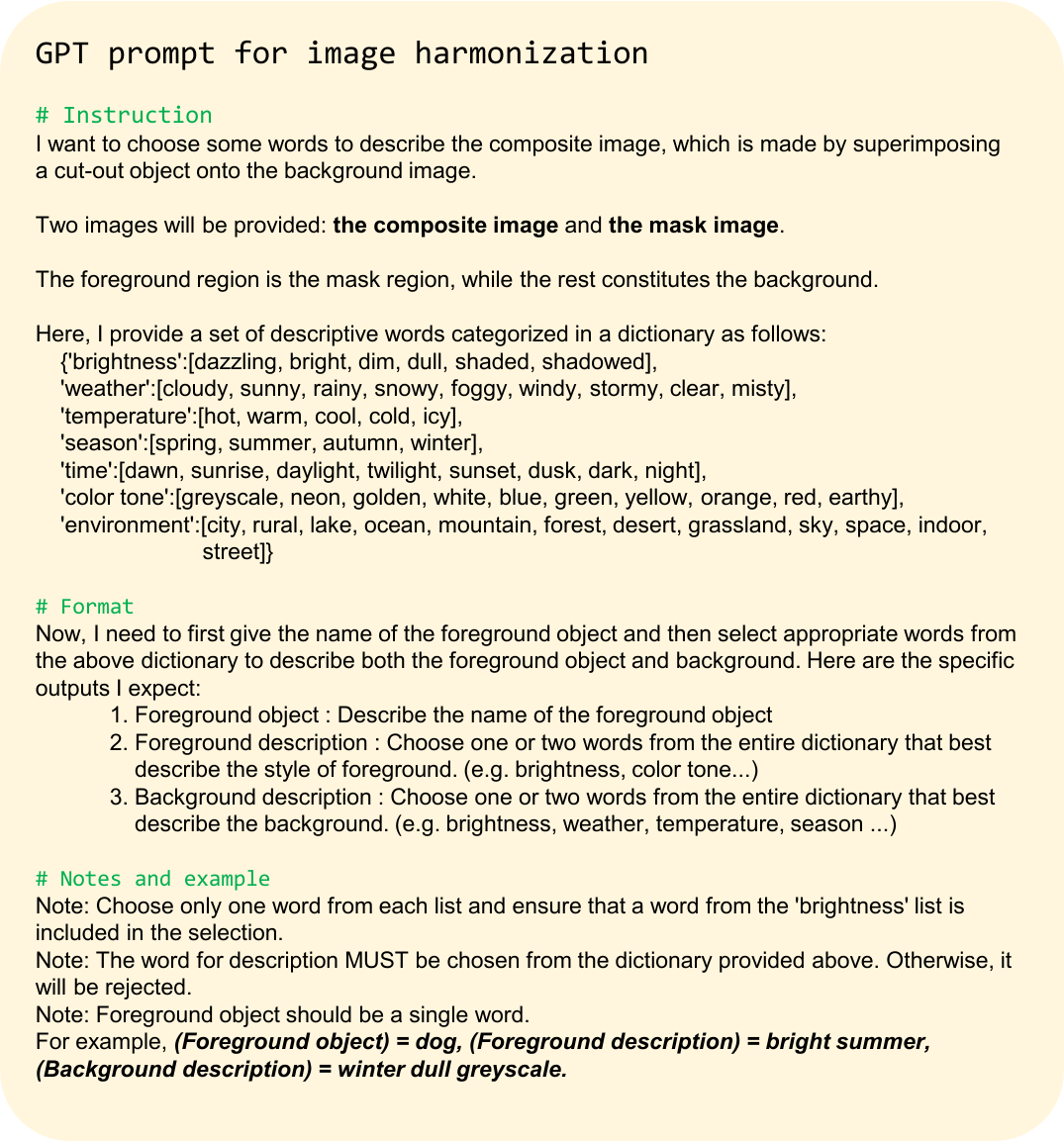}
\end{center}
\caption{
A prompt for using GPT-4o~\cite{gpt4} as a prompt generator for image harmonization.
}
\label{fig:gpt_prompt_image_harmonization}
\end{figure*}

%% file: figures/gpt_prompt_style_transfer.tex
\begin{figure*}[t!] 
\begin{center}
\includegraphics[width=\linewidth]{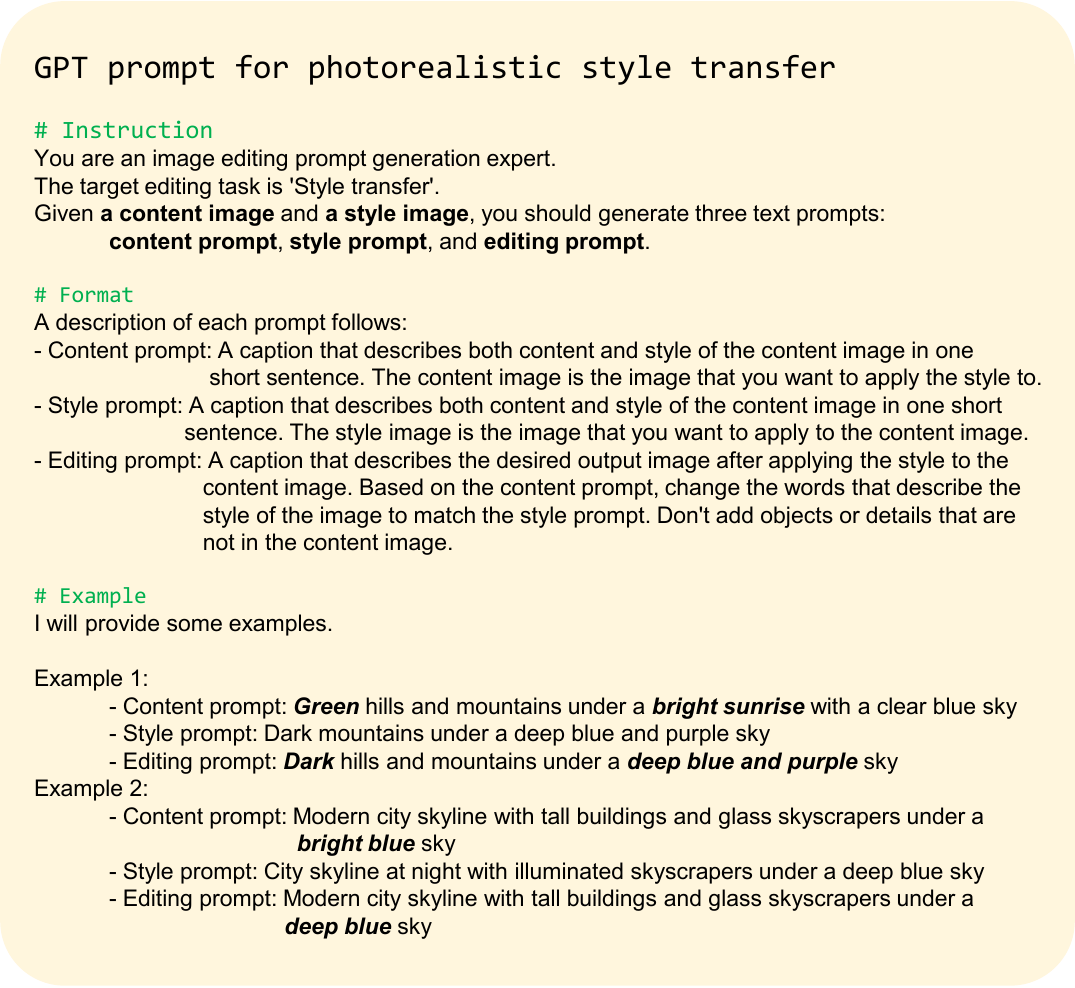}
\end{center}
\caption{
A prompt for using GPT-4o~\cite{gpt4} as a prompt generator for photorealistic style transfer.
}
\label{fig:gpt_prompt_style_transfer}
\end{figure*}